\documentclass[runningheads]{llncs}

\usepackage{eccv}

\usepackage{eccvabbrv}

\usepackage{graphicx}
\usepackage{booktabs}

\usepackage{url}
\usepackage{tabularx} 
\usepackage{caption}
\usepackage[table]{xcolor}
\usepackage{algorithm}
\usepackage{algorithmic}
\usepackage{float}
\usepackage{enumitem}
\usepackage{multirow}
\usepackage{xcolor}
\usepackage[table]{xcolor}
\usepackage[skins,breakable]{tcolorbox}

\definecolor{momapink}{HTML}{FFF0F5}  
\definecolor{momayellow}{HTML}{FFFFE0} 

\definecolor{aliceblue}{HTML}{F0F8FF}  
\definecolor{honeydew}{HTML}{F0FFF0}  

\newcommand{\ours}{RAG-3DSG}

\usepackage[accsupp]{axessibility}  

\usepackage{hyperref}

\usepackage{orcidlink}

\begin{document}

\title{RAG-3DSG: Enhancing 3D Scene Graphs with Re-Shot Guided Retrieval-Augmented Generation}

\titlerunning{RAG-3DSG}

\author{Yue Chang\inst{1} \and
Rufeng Chen\inst{1} \and
Zhaofan Zhang\inst{1} \and\\
Yi Chen\inst{2} \and
Yifan Tian\inst{1} \and
Sihong Xie\inst{1}\thanks{Corresponding author.}}

\authorrunning{Y.~Chang et al.}

\institute{The Hong Kong University of Science and Technology (Guangzhou), China \and
Jilin University, China\\
\email{ychang500@connect.hkust-gz.edu.cn}, \email{sihongxie@hkust-gz.edu.cn}}

\maketitle

\begin{abstract}
Open-vocabulary 3D Scene Graph (3DSG) can enhance various downstream tasks in robotics by leveraging structured semantic representations, yet current 3DSG construction methods suffer from semantic inconsistencies caused by noisy cross-image aggregation under occlusions and constrained viewpoints. 
To mitigate the impact of such inconsistency, we propose \textbf{\ours}, which introduces re-shot guided uncertainty estimation. By measuring the semantic consistency between original limited viewpoints and re-shot optimal viewpoints, this method quantifies the underlying semantic ambiguity of each graph object.
Based on this quantification, we devise an Object-level Retrieval-Augmented Generation (RAG) that leverages low-uncertainty objects as semantic anchors to retrieve more reliable contextual knowledge, enabling a Vision-Language Model to rectify the predictions of uncertain objects and optimize the final 3DSG.
Extensive evaluations across three challenging benchmarks and real-world robot trials demonstrate that \ours \space achieves superior recall and precision, effectively mitigating semantic noise to provide more reliable scene representations for robotics tasks.
\end{abstract}

\section{INTRODUCTION}
Compact and expressive representation of complex and semantic-rich 3D scenes has long been a fundamental challenge in robotics, with direct impact on downstream tasks such as robot manipulation \cite{shridhar2022cliport, rashid2023language, honerkamp2024language, yan2025dynamic} and navigation \cite{gadre2022clip, shah2023lm, yin2024sg, werby2024hierarchical, yan2025dynamic, chen2026psg}. 
A promising representation is 3D scene graphs~(3DSGs) \cite{armeni20193d, gay2018visual}, which encode a scene into a graph where nodes denote objects and edges capture their pairwise relationships.
Early efforts focus on building 3DSGs by detecting objects and relationships from a closed  vocabulary \cite{hughes2022hydra, rosinol2021kimera, wu2021scenegraphfusion}. 
While these methods perform well and are efficient in fixed environments, their reliance on a closed vocabulary restricts their generalization to novel, unseen scenes in the complex real-world.
To mitigate this limitation, recent approaches \cite{gu2024conceptgraphs, werby2024hierarchical, koch2024open3dsg, maggio2024clio, jatavallabhula2023conceptfusion, linok2025beyond, fungraph3d, yan2025dynamic, yamazaki2024open} leverage vision-language foundation models to provide open-vocabulary 3DSG generation and produce more expressive representations for diverse scenes.

\begin{figure}[!b]
\begin{center}
\includegraphics[width=\linewidth]{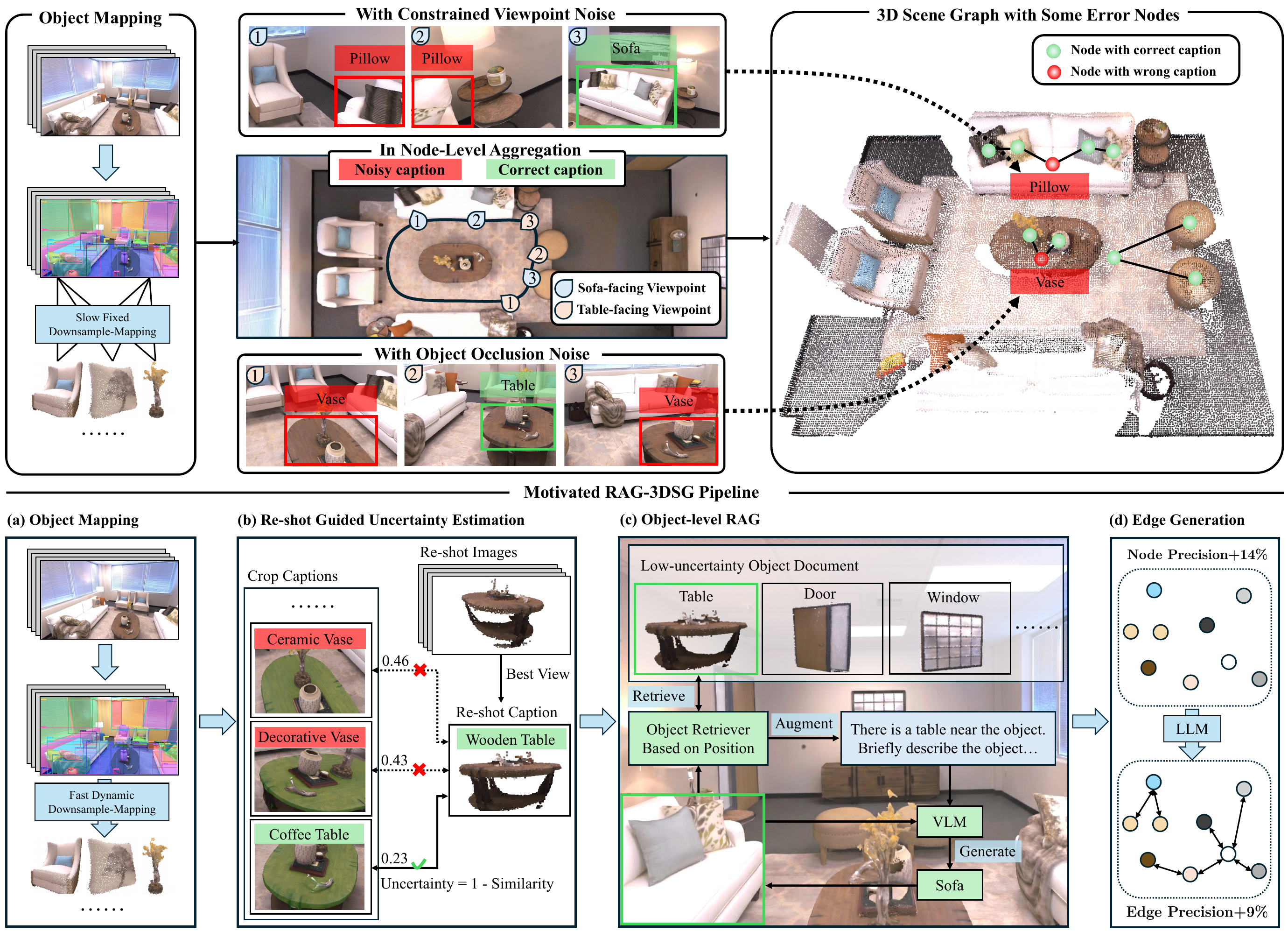}
\end{center}
\caption{
An overview of the \ours \space framework.
The upper part illustrates common challenges in multi-view 3D scene graph generation.
Our pipeline addresses these issues through
\textbf{(a)} Multi-view RGB-D frames are segmented and fused into a global object list with point clouds and semantic embeddings. 
\textbf{(b)} Re-shot images are used to select the best-view caption, which is compared with crop captions to estimate uncertainty; clustering is applied and the top-1 cluster is retained for fusion. 
\textbf{(c)} Low-uncertainty objects form a retrieval document, while high-uncertainty objects leverage retrieved context for caption refinement via a VLM. 
\textbf{(d)} Finally, spatial and semantic relationships among objects are inferred by an LLM to construct the 3D scene graph.
}
\label{fig:structure}
\end{figure}

Despite this progress, open-vocabulary approaches largely adhere to the one-way pipeline of per-image object-level information extraction followed by cross-image aggregation.
However, as illustrated in the upper part of~\cref{fig:structure}, constrained viewpoints, occlusions, and other poor imaging conditions can introduce a significant amount of noise, leading to semantic inconsistency where the same object is described inconsistently across different views. 
For example, as shown in~\cref{fig:structure} (b), although the object of interest is the \textit{table}, the presence of an occluding \textit{vase} leads to multiple crop captions being mistakenly recognized as \textit{vase}.
Crucially, existing methods often blindly aggregate these conflicting captions or embeddings across images without discrimination, allowing high-noise observations to corrupt the accuracy of the resulting 3DSG. Such noise-induced inaccuracies in 3DSGs are unacceptable for downstream robotic tasks, particularly in safety-critical scenarios. 
For example, if a medication bottle is incorrectly described as a beverage container, a service robot could deliver dangerous substances to humans.
Therefore, rather than blindly trusting aggregated features, we argue that the system must "know what it doesn't know"—quantifying semantic ambiguity to distinguish reliable predictions from inconsistent ones.

To this end, we propose \textbf{\ours}, a framework that mitigates aggregation noise through a two-phase "diagnose-then-rectify" paradigm. 
The core challenge lies in breaking the dependency on finite input trajectories; while existing methods \cite{gorlo2025describe, linok2025beyond, ju2025momagraph} filter for the best available object frames, they remain "passive observers" restricted to captured viewpoints—even when the true clear object was never recorded. 
To transform this limited observation into active verification \cite{vazquez2001viewpoint, zhao2026cov, zhang2026think3d}, our first phase conceptually shifts the system's role to an \textit{active photographer}. 
Since reconstructed 3D geometry naturally disentangles objects from background clutter, we virtually `re-shoot' objects from optimal angles. 
By comparing captions generated from these unconstrained re-shots against the original passive crops (\cref{fig:structure} (b)), we quantify the semantic ambiguity of each object. 
However, identifying uncertainty is only half the battle; rectifying it without hallucination requires robust context. 
Therefore, in the second phase, we use these uncertainty scores to drive our Object-level Retrieval-Augmented Generation (RAG) module. 
Instead of forcing the Vision-Language Model (VLM) to guess semantics in isolation, low-uncertainty objects are anchored as reliable context. 
High-uncertainty objects then trigger the retrieval of surrounding low-uncertainty "semantic anchors" on the 3DSG to guide object caption refinement (\cref{fig:structure} (c)), which iteratively improves the overall 3DSG (\cref{fig:structure} (d)).

To validate the effectiveness of our approach, we conduct extensive experiments on two challenging real-world benchmarks, SceneFun3D \cite{scenefun3d} and FunGraph3D \cite{fungraph3d}, alongside a rigorous fine-grained human evaluation on the Replica \cite{straub2019replica} dataset. 
Quantitative results demonstrate that \ours \space establishes a new state-of-the-art in both object classification and relationship prediction. 
Furthermore, human evaluation confirms that our method yields significantly cleaner and more precise scene graphs, substantially suppressing duplicate node predictions and relational hallucinations. 
Finally, robust performance under injected tracking noise on real-world datasets, coupled with successful real-world robotic experiments, demonstrates that our active re-shot mechanism effectively withstands noise, underscoring its readiness for reliable real-world deployment.

\section{RELATED WORK}
\textbf{Scene Graph Generation (SGG)}
Scene graphs were initially introduced in the 2D domain to extract objects and their relationships from images, providing a structured representation that supports a highly abstract understanding of scenes for intelligent agents \cite{sun2023evidential, liu2021fully, yin2018zoom, krishna2017visual, lu2016visual, johnson2015image}.
A scene graph is typically composed of three fundamental elements—objects, attributes, and relations—and can be expressed as a set of visual relationship triplets in the form of \textless subject, relation, object\textgreater \cite{li2024scene}.
In the field of 3D scene representations, several works \cite{wald2020learning, kim20193, armeni20193d} have drawn inspiration from 2D scene graphs and extended the concept into the 3D domain. 
Compared to dense, hard-to-interpret point clouds with per-point semantics, 3D scene graphs (3DSGs) provide a more compact and structured abstraction of the scene. 
By organizing objects and their relationships into a graph representation, they enable more efficient reasoning and facilitate downstream tasks such as robotic navigation~\cite{gadre2022clip, shah2023lm, yin2024sg, werby2024hierarchical, yan2025dynamic, chen2026psg} and scene understanding \cite{rana2023sayplan, agia2022taskography}.
Early efforts in 3D scene graph generation~(3DSGG) enabled the construction of real-time systems capable of dynamically building hierarchical 3D scene representations \cite{hughes2022hydra, rosinol2021kimera, wu2021scenegraphfusion}. 
However, these methods were strictly confined to the closed-vocabulary setting, which restricted their applicability to a limited range of tasks.
More recently, several works \cite{gu2024conceptgraphs, werby2024hierarchical, koch2024open3dsg, maggio2024clio, jatavallabhula2023conceptfusion, linok2025beyond, fungraph3d, yan2025dynamic, yamazaki2024open} have begun to explore open-vocabulary approaches for 3D scene graph generation. 
Using Vision Language Models (VLMs) and Large Language Models (LLMs), these methods sacrifice part of the real-time capability but significantly expand the range of object categories and relations that can be recognized, thus broadening the applicability to a wider variety of complex downstream tasks.

\textbf{Open-vocabulary 3DSGG}
Recent advances have extended 3DSGG to the open-vocabulary setting by leveraging VLMs and LLMs. 
Existing approaches can be divided into two main paradigms.
The first paradigm follows a caption-first strategy, where objects in each image are independently described and later aggregated into scene-level semantics. 
In practice, multiple masked views of the same object are captioned separately, and a LLM is then used to aggregate these descriptions into a final caption (e.g., \cite{gu2024conceptgraphs}).
The second paradigm~\cite{werby2024hierarchical, koch2024open3dsg, jatavallabhula2023conceptfusion, yamazaki2024open} adopts an embedding-first strategy, where semantic embeddings of multiple masked views of the same object are first extracted and aggregated without explicit captioning. 
The aggregated embeddings are then converted into captions or directly aligned with user queries in a joint vision–language space using models such as CLIP~\cite{radford2021learning} or SEEM~\cite{zou2023segment}.
In addition, some works assign captions to object embeddings by matching them against an arbitrary vocabulary with CLIP, a strategy commonly referred to as open-set 3DSG~\cite{maggio2024clio}.
Despite their differences, both paradigms share the same underlying pipeline of per-image information extraction followed by cross-image information aggregation. 
However, the information extraction process is often affected by constrained viewpoints and object occlusion, which introduce noise into the representations. 
As a result, aggregated information may be inaccurate, leading to reduced precision in the node and edge captions of the 3DSGs.
This limitation cannot be simply resolved by adopting more powerful captioning models, as the noise originates from inherent challenges in multi-view perception such as occlusion and viewpoint constraints.

\textbf{Viewpoint-Aware 3D Semantic Reasoning} The quality of semantic understanding is intrinsically linked to observation viewpoints~\cite{vazquez2001viewpoint}. To mitigate occlusion in complex scenes, recent 3DSG methods~\cite{gorlo2025describe, linok2025beyond, ju2025momagraph} have adopted frame-selection strategies to ensure robust object captioning. For instance, BBQ~\cite{linok2025beyond} clusters camera poses to select the frame with the maximum projected area for description generation, while MomaGraph~\cite{ju2025momagraph} filters the input stream for significant keyframes to obtain informative visual context. Despite their effectiveness, these approaches remain passive: they are strictly bound by the finite input trajectory, selecting only the "best of the available" views even if the optimal angle was never captured.
In contrast, emerging research in the broader field of general 3D perception advocates active spatial reasoning using 3D geometry. Think3D~\cite{zhang2026think3d} proposes an interactive framework where agents actively manipulate viewpoints, transforming spatial reasoning into a "3D Chain-of-Thought" process. Similarly, Chain-of-View (CoV)~\cite{zhao2026cov} converts VLMs into active viewpoint reasoners, employing a coarse-to-fine exploration strategy to expand "finite observations" into an "infinite" verification space. 
Our \ours{} bridges this gap by introducing active verification into open-vocabulary 3DSG generation. 
\section{METHOD}
In this section, we present \ours, a framework designed to bridge the gap between passive perception and active verification in open-vocabulary 3D scene graph generation.
As illustrated in~\cref{fig:structure}, our pipeline consists of three main stages:
(1) \textbf{Cross-image Object Mapping} (\cref{subsec:inter_frame_object_mapping}), 
where we incrementally fuse local observations into a global object list, employing dynamic downsampling to balance representation efficiency with point cloud density;
(2) \textbf{Node Caption Generation} (\cref{subsec:node_caption_generation}), 
which implements our diagnose-then-rectify mechanism: utilizing re-shot guided uncertainty estimation to identify semantic ambiguity, followed by Object-level RAG to rectify unreliable predictions;
(3) \textbf{Edge Caption Generation} (\cref{subsec:edge_caption_generation}), 
where we leverage the refined object nodes to prompt an LLM for interpretable inter-object relationship generation.

\subsection{Cross-image Object Mapping}
\label{subsec:inter_frame_object_mapping}
Given an RGB-D image sequence $I = \{I_1, \dots, I_t\}$ with poses $P = \{P_1, \dots, P_t\}$, we maintain a global object list $O^{\text{global}}$ by incrementally fusing local objects. 
This process comprises local information extraction followed by global fusion.

\subsubsection{Local Information Extraction}
For each incoming frame $I_t$, we first extract local object instances. 
A class-agnostic segmentation model $\text{SAM}(\cdot)$~\cite{kirillov2023segment} generates 2D masks $\{m_{t,i}\}$, which are then encoded by $\text{CLIP}(\cdot)$~\cite{radford2021learning} to obtain semantic embeddings $\{f_{t,i}\}$. 
Using the camera pose $P_t$, we project these masks into 3D space to form point clouds $\{p_{t,i}\}$. 
This yields a local object list $O_t^{\text{local}} = \{ \langle f_{t,i}^{\text{local}}, p_{t,i}^{\text{local}} \rangle \}_{i=1}^M$, serving as the raw input for global fusion.

\subsubsection{Global Fusion with Dynamic Downsampling}
\label{subsec:3d_object_fusion}
To integrate $O_t^{\text{local}}$ into the global list $O_{t-1}^{\text{global}}$, we employ an incremental fusion pipeline.
However, standard fusion with fixed-size voxelization often leads to a trade-off: it either over-samples large background structures or under-samples small objects (losing details crucial for our subsequent re-shot verification).
To address this, we apply a scale-adaptive downsampling strategy prior to fusion. 
The voxel size $\delta_{t,i}^{\text{voxel}}$ for the i-th object in the t-th frame is dynamically adjusted based on its spatial extent:
\begin{equation}
\label{eq:voxel_size}
    \delta_{t,i}^{\text{voxel}} = \delta^{\text{sample}} \cdot \|\text{Bbox}(p_{t,i})\|_2^{1/2},
\end{equation}
where $\delta^{\text{sample}}$ is a base constant and $\|\text{Bbox}(p_{t,i})\|_2$ is the diagonal length of the object's 3D bounding box.
This strategy ensures that small objects retain high point density for reliable re-identification, while large objects remain efficient.

Following dynamic downsampling, we associate local object $o_{t,i}^{\text{local}}$ with global object $o_{t-1,j}^{\text{global}}$ by maximizing a composite similarity score $\theta = \theta_{\text{sem}} + \theta_{\text{spa}}$.
The semantic similarity $\theta_{\text{sem}}$ is the cosine similarity between the CLIP embeddings. 
Spatial similarity $\theta_{\text{spa}}$ is defined as the \textit{dynamic nearest neighbor ratio} to measure point cloud overlap. Rather than a fixed radius, its matching threshold scales dynamically with object sizes (derived from \cref{eq:voxel_size}).
Pairs exceeding a similarity threshold $\delta^{\text{sim}}$ are merged: the point clouds are unified ($p^{\text{global}} \leftarrow p^{\text{global}} \cup p^{\text{local}}$), and embeddings are updated via a moving average to accumulate semantic information. Unmatched local objects are registered as new global instances.

\subsection{Node Caption Generation}
\label{subsec:node_caption_generation}
Having constructed the global object list, we now enter the core active verification phase.
Instead of passively accepting the noisy semantics from input frames, we implement a diagnose-then-rectify mechanism.
As shown in \cref{fig:structure}, we first derive initial captions from passive observations, and then introduce a re-shot guided uncertainty estimation module.
By actively synthesizing ``best-view'' re-shots to verify the initial captions, we rigorously quantify semantic ambiguity, separating reliable anchors from uncertain nodes.
These verified anchors serve as the knowledge base for our Object-level RAG, which retrieves spatial context to rectify the ambiguous predictions, ensuring globally consistent node semantics.

\subsubsection{Initial Caption Generation: The Passive Baseline}
\label{subsec:initial_caption_generation}
For each global object $o_{t,i}^{\text{global}}$, we first extract semantic information from the original input sequence.
We maintain the top-$k$ views with the highest segmentation confidence and feed these object-level crops into a VLM \cite{hurst2024gpt} to obtain initial captions $c_{t,i} = \{ c_1, \dots, c_k \}$.
However, as illustrated in \cref{fig:structure} (Top), this process is inherently passive: it is strictly bound by the camera's trajectory.
If the camera only captures an object from a constrained or occluded viewpoint, the resulting captions $c_{t,i}$ will inevitably be noisy or incorrect, necessitating an active verification mechanism.

\subsubsection{Re-shot Guided Uncertainty Estimation: Active Verification}
\label{subsec:reshot_guided_uncertainty_evaluation}
To break the dependency on passive input views, we propose an active re-shot strategy.
Conceptually, we transform the role of the observer from a passive recorder to an active photographer.
Since the reconstructed point cloud $p$ aggregates temporal observations into a holistic geometry, it is disentangled from transient occlusion and background clutter, offering an unconstrained verification space.
Our goal is to simulate an active optimal photographer that "re-shoots" the object from a scientifically determined "best view" to capture its underlying semantics.

\begin{figure}[ht]
\begin{center}
\includegraphics[width=\linewidth]{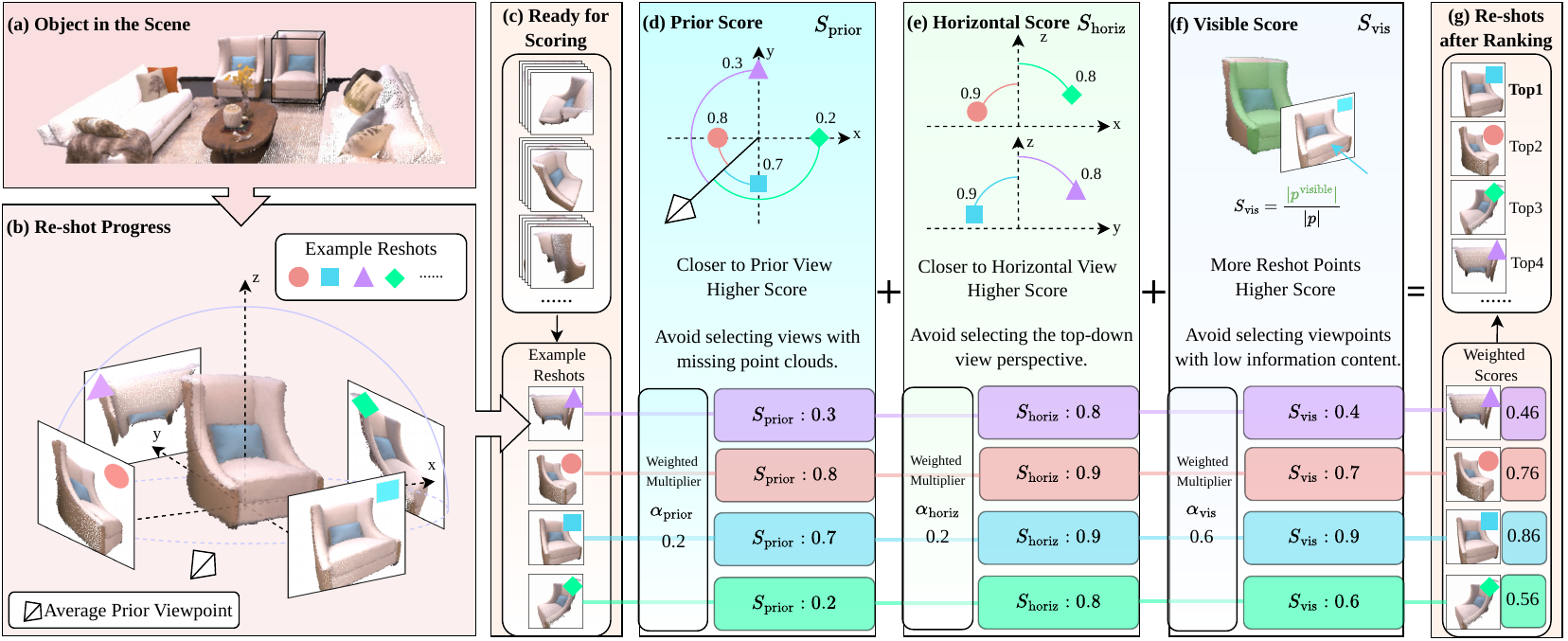}
\end{center}
\caption{
Workflow of the Re-shot Guided Uncertainty Estimation.
\textbf{(a-b)} We reconstruct the object and simulate active re-shots on a sampling hemisphere.
\textbf{(c-g)} A robust multi-factor scoring system ($S_\text{prior}, S_\text{horiz}, S_\text{vis}$) evaluates each candidate view to identify the most informative angle, effectively avoiding ambiguity caused by extreme perspectives.
}
\label{fig:pic2}
\end{figure}

As depicted in \cref{fig:pic2} (b), we uniformly sample 64 virtual candidate cameras on a hemisphere centered at the target object $o$.
To identify the most informative view among these candidates, we design a robust view quality scoring system that mimics human preference for clear visual observation.
As shown in \cref{fig:pic2} (d-f), the score for a candidate view $c_i$ comprises three complementary terms:
\begin{equation}
S_\text{vis} = \frac{|p^\text{visible}|}{|p|}, \quad
S_\text{horiz} = 1 - |{v}_i \cdot {g}|, \quad
S_\text{prior} = \tfrac{1}{2}(1 + {v}_i \cdot {f}).
\label{eq:view_scores}
\end{equation}
Specifically, $S_\text{vis}$ maximizes information content by preferring views with high point visibility;
$S_\text{horiz}$ avoids disorientation by penalizing top-down or bottom-up views that are aligned with the gravity vector $g$ (i.e., the downward z-axis in the gravity-aligned world frame);
and $S_\text{prior}$ maintains consistency with the average prior viewpoint $f$ (the average viewing direction of the source frames observing the object) to avoid complete hallucinations from unseen backsides.
The weighted sum $S_{\text{view}} = \alpha_{\text{vis}} S_{\text{vis}} + \alpha_{\text{horiz}} S_{\text{horiz}} + \alpha_{\text{prior}} S_{\text{prior}}$ (subject to $\sum \alpha = 1$) guides us to render the optimal re-shot image $I^{\text{reshot}}$ (see \cref{fig:pic2} (g)).
Importantly, since candidate scoring relies purely on lightweight operations and only the single top-scoring view triggers rendering and VLM querying, this active verification introduces minimal computational overhead.

With the optimal re-shot caption $c^{\text{reshot}}$ generated from $I^{\text{reshot}}$, we quantify the object reliability by comparing this "active verification" against the "passive observations".
We compute the cosine similarities between their  embeddings:
\begin{equation}
\{ s_1, \dots, s_k \} = \left\{ \cos \big( \text{CLIP}(c^{\text{reshot}}), \text{CLIP}(c_i) \big) \right\}_{i=1}^k.
\end{equation}
High similarity implies that the initial captions align with the reshot one, indicating high reliability.
Conversely, low similarity signals conflict, flagging the object as uncertain.
Specifically, these similarity scores are clustered using the KMeans algorithm with $K=3$.
The cluster exhibiting the highest mean similarity is designated as the consensus cluster.
The initial passive captions within this consensus cluster, together with the active re-shot caption $c^{\text{reshot}}$, are then aggregated by an LLM to generate the final consensus caption $\hat{c}$.
Accordingly, the average similarity score of this consensus cluster is derived as the object's reliability score $\hat{s}$, which serves as the criterion for the subsequent RAG.

\subsubsection{Object-level RAG: Contextual Rectification}
\label{subsec:object_level_RAG_construction}
Following the uncertainty diagnosis, we proceed to the rectification phase.
We first rank all objects by their uncertainty scores $1 - \hat{s}$ and apply a prompt to the VLM~\cite{hurst2024gpt} to filter out background objects via crops. 
The top-$50\%$ low-uncertainty objects are included in the object document for RAG, where each final caption $c$ is set to $\hat{c}$.
For the remaining high-uncertainty objects, we perform refinement using contextual information.
Specifically, a 3D position-based retriever retrieves the nearest object in the document, whose caption $c^{\text{env}}$ serves as augmented auxiliary context.
In addition, we construct a composite image by concatenating the re-shot image (providing global context) with the crop image that yields the highest similarity score.
This composite image, together with a text prompt containing $c^{\text{env}}$, is fed into a VLM~\cite{hurst2024gpt} to generate the refined caption $c$.
The prompt is designed as:  
``The picture is stitched from the point cloud image and the RGB image of the same indoor object. There is a $c^{\text{env}}$ near the object. Briefly describe the object in the picture.''
Through the refinement process, we obtain a precise object list 
$O = \{ o_{1}, \dots , o_{N} \}$, 
where each object $o_i$ is represented as 
$o_{i} = \langle f_{i}, p_{i}, c_{i} \rangle$, 
consisting of semantic embedding $f_i$, pointcloud $p_i$, and precise node caption $c_i$.

\subsection{Edge Caption Generation}
\label{subsec:edge_caption_generation}
Following \cite{gu2024conceptgraphs}, we estimate spatial relationships among 3D objects to complete the scene graph.
For connectivity, we diverge from the fixed NN-ratio of \cite{gu2024conceptgraphs} by introducing a dynamic threshold aligned with our downsampling strategy (\cref{eq:voxel_size}), adapting edge construction to varying point densities.
For relationship captioning, although our paradigm inherently supports an open-vocabulary relation space, we intentionally constrain the final taxonomy to eight categories. These comprise bi-directional pairs for \textit{support} (on), \textit{containment} (in), and \textit{part-whole} (part of), alongside \textit{proximity} (near) and \textit{none}.
This multi-class constraint is strategically introduced to eliminate synonym ambiguity and ensure the stability of the LLM's spatial reasoning patterns.
Moreover, we incorporate few-shot in-context examples.
This design provides the model with explicit spatial reasoning patterns, ensuring that the generated edge captions are more expressive and accurate, thereby establishing a more robust structured representation for complex downstream robotics applications.
\section{EXPERIMENTS}
\subsection{Experimental Setup}

\noindent\textbf{Datasets \& Benchmarks.} 
We evaluate our method on two real-world datasets: \textbf{SceneFun3D}~\cite{scenefun3d} (20 scenes) and \textbf{FunGraph3D}~\cite{fungraph3d} (24 scenes), strictly following the OpenFunGraph~\cite{fungraph3d} protocol. These datasets feature real-world challenges like sensor noise and clutter, inherently distinct from synthetic data. To align with standard 3DSSG definitions~\cite{wald2020learning}, we filter ground truths to retain only object nodes and semantic edges, excluding specialized functional interaction elements. Additionally, we use \textbf{Replica}~\cite{straub2019replica} exclusively for fine-grained human evaluation, leveraging its clean geometry for unambiguous semantic assessment.

\noindent\textbf{Baselines \& Implementation Variants.}
We compare against SOTA open-vocabulary methods: \textbf{Open3DSG}~\cite{koch2024open3dsg}, \textbf{ConceptGraphs}~\cite{gu2024conceptgraphs} (and its Detector variant), and \textbf{OpenFunGraph}~\cite{fungraph3d}. To demonstrate scalability, we evaluate two framework variants: \textbf{Ours-LLaVA} (using a local LLaVA-v1.5-7b~\cite{liu2023llava} for fair comparison) and \textbf{Ours-GPT} (using GPT-4o~\cite{hurst2024gpt} to probe the performance upper bound).  
Additionally, three compact settings are included for mechanism analysis: 
(i) \textbf{Passive VLM} completely bypasses active re-shooting and aggregates captions solely from the top-3 highest-confidence passive crops. 
(ii) \textbf{Re-shot VLM} strips away the RAG module and image concatenation, directly adopting the isolated re-shot caption as the final node semantic. 
(iii) \textbf{Ours Node + CG Edge} pairs our rectified nodes with the vanilla ConceptGraphs~\cite{gu2024conceptgraphs} edge generator to isolate node-level contributions.

\noindent\textbf{Implementation Details.}
We utilize SAM (sam\_vit\_h\_4b8939) and CLIP (ViT-H-14) for segmentation and semantic extraction. Reasoning is powered by GPT-4o or LLaVA. Key hyperparameters include: similarity threshold $\delta_{sim} = 0.45$ (balancing over-segmentation and distinct instance discovery), base voxel size $\delta_{sample} = 0.01$m, and balanced scoring weights ($\alpha_{prior} = \alpha_{horiz} = 0.2, \alpha_{vis} = 0.6$) for uncertainty estimation. Exhaustive parameter analyses, prompt templates, and runtime profiling are deferred to the Supplementary Material.

\noindent\textbf{Robustness Setting.}
To address concerns that 3D-centric pipelines over-rely on perfect multi-view reconstructions, we introduce an extreme stress test. 
We deliberately inject Gaussian noise ($\sigma_{trans} = 0.02$\,m, $\sigma_{rot} = 1^\circ$) into camera extrinsics. 
This \textit{w/ Noisy Pose} setting (\cref{tab:main_results}) empirically verifies if our re-shot and RAG mechanisms remain effective despite severely degraded point clouds.

\noindent\textbf{Evaluation Metrics.}
For our main evaluation on SceneFun3D and FunGraph3D, we report Recall@K for both \textit{Objects} (node classification) and \textit{Edges} (predicate prediction). 
To ensure strictly fair comparisons and handle open-vocabulary edge predictions, we directly adopt the evaluation protocol of OpenFunGraph~\cite{fungraph3d}: predicted predicates are mapped to ground-truth relationships based on the cosine similarity of their BERT embeddings, preventing any mapping bias from our expanded relation space. 
For the fine-grained Human Evaluation on Replica~\cite{straub2019replica}, three independent experts assess semantic correctness using a strict annotation protocol, with inter-annotator agreement verified via Fleiss’ Kappa ($\kappa$).

\subsection{Main Results}
\label{sec:main_results}

\noindent\textbf{1) Object Classification (Node).} 
Our method establishes a new state-of-the-art for object recognition across both datasets. 
As shown on the SceneFun3D split, \textbf{Ours-GPT} achieves a remarkable 90.2\% R@10, outperforming the previous best method, OpenFunGraph (87.8\%). 
This dominance extends to the more cluttered and challenging FunGraph3D benchmark, where \textbf{Ours-GPT} reaches 75.0\% in R@3 and 83.0\% in R@10, exceeding the SOTA by substantial margins of 4.3\% and 3.9\%, respectively. 
Crucially, even our locally deployed open-weight model (\textbf{Ours-LLaVA}) surpasses OpenFunGraph (73.6\% vs. 70.7\% in R@3). 
This compelling result underscores that our \textit{diagnose-then-rectify} framework effectively compensates for raw model capacity, enabling accessible VLMs to achieve top-tier perception through active verification and RAG.

\begin{table*}[t]
\centering
\caption{
\textbf{Quantitative comparison of 3DSG generation on real-world datasets.} 
Following the standard evaluation protocol, we report Recall@K (R@K) for both Object classification (Nodes) and Relationship prediction (Edges) on the SceneFun3D (\colorbox{momapink}{pink columns}) and FunGraph3D (\colorbox{momayellow}{yellow columns}). 
\textbf{Bold} and \underline{underline} denote the best and second-best performances, respectively. 
$^\ast$ indicates the usage of OpenFunGraph's fused 3D nodes rather than the ground-truth for a fair comparison.
}
\label{tab:main_results}

\renewcommand{\arraystretch}{1.25} 
\setlength{\tabcolsep}{5pt}       

\resizebox{\textwidth}{!}{%
\begin{tabular}{cccccccccc}
\toprule
\multirow{2}{*}{\textbf{Type}} & \multirow{2}{*}{\textbf{Method}} & \multicolumn{4}{c}{\cellcolor{momapink}\textbf{SceneFun3D}} & \multicolumn{4}{c}{\cellcolor{momayellow}\textbf{FunGraph3D}} \\
\cmidrule(lr){3-6} \cmidrule(lr){7-10}
 & & \multicolumn{2}{c}{\cellcolor{momapink}Objects} & \multicolumn{2}{c}{\cellcolor{momapink}Edges} & \multicolumn{2}{c}{\cellcolor{momayellow}Objects} & \multicolumn{2}{c}{\cellcolor{momayellow}Edges} \\
 & & \cellcolor{momapink}R@3 & \cellcolor{momapink}R@10 & \cellcolor{momapink}R@5 & \cellcolor{momapink}R@10 & \cellcolor{momayellow}R@3 & \cellcolor{momayellow}R@10 & \cellcolor{momayellow}R@5 & \cellcolor{momayellow}R@10 \\
\midrule

\multirow{5}{*}{\rotatebox{90}{\textbf{Baselines}}} 
 & Open3DSG~\cite{koch2024open3dsg} & \cellcolor{momapink}61.2 & \cellcolor{momapink}70.7 & \cellcolor{momapink}69.2 & \cellcolor{momapink}78.8 & \cellcolor{momayellow}47.6 & \cellcolor{momayellow}55.7 & \cellcolor{momayellow}47.9 & \cellcolor{momayellow}55.9 \\
 & Open3DSG$^\ast$~\cite{koch2024open3dsg} & \cellcolor{momapink}42.9 & \cellcolor{momapink}50.0 & \cellcolor{momapink}64.4 & \cellcolor{momapink}72.3 & \cellcolor{momayellow}30.9 & \cellcolor{momayellow}44.1 & \cellcolor{momayellow}46.6 & \cellcolor{momayellow}55.7 \\
 & ConceptGraphs~\cite{gu2024conceptgraphs} & \cellcolor{momapink}71.3 & \cellcolor{momapink}77.1 & \cellcolor{momapink}80.2 & \cellcolor{momapink}95.0 & \cellcolor{momayellow}56.6 & \cellcolor{momayellow}65.6 & \cellcolor{momayellow}51.5 & \cellcolor{momayellow}84.6 \\
 & ConceptGraphs (Det)~\cite{gu2024conceptgraphs} & \cellcolor{momapink}72.3 & \cellcolor{momapink}78.6 & \cellcolor{momapink}81.3 & \cellcolor{momapink}92.0 & \cellcolor{momayellow}57.5 & \cellcolor{momayellow}68.0 & \cellcolor{momayellow}52.4 & \cellcolor{momayellow}80.2 \\
 & OpenFunGraph~\cite{fungraph3d} & \cellcolor{momapink}\underline{81.8} & \cellcolor{momapink}87.8 & \cellcolor{momapink}\textbf{88.1} & \cellcolor{momapink}\underline{96.2} & \cellcolor{momayellow}70.7 & \cellcolor{momayellow}79.1 & \cellcolor{momayellow}\underline{65.1} & \cellcolor{momayellow}\textbf{91.4} \\
\midrule

\multirow{4}{*}{\rotatebox{90}{\textbf{Analysis}}} 
 & Ours Node + CG Edge & \cellcolor{momapink}-- & \cellcolor{momapink}-- & \cellcolor{momapink}84.1 & \cellcolor{momapink}95.6 & \cellcolor{momayellow}-- & \cellcolor{momayellow}-- & \cellcolor{momayellow}61.2 & \cellcolor{momayellow}89.4 \\
 & Passive VLM (gpt-4o) & \cellcolor{momapink}67.7 & \cellcolor{momapink}72.9 & \cellcolor{momapink}78.3 & \cellcolor{momapink}86.3 & \cellcolor{momayellow}53.4 & \cellcolor{momayellow}66.1 & \cellcolor{momayellow}49.5 & \cellcolor{momayellow}78.2 \\
 & Passive VLM (gpt-5.2) & \cellcolor{momapink}78.9 & \cellcolor{momapink}85.6 & \cellcolor{momapink}80.6 & \cellcolor{momapink}89.1 & \cellcolor{momayellow}70.4 & \cellcolor{momayellow}78.6 & \cellcolor{momayellow}60.6 & \cellcolor{momayellow}84.1 \\
 & Re-shot VLM (gpt-4o) & \cellcolor{momapink}59.4 & \cellcolor{momapink}72.3 & \cellcolor{momapink}76.7 & \cellcolor{momapink}85.5 & \cellcolor{momayellow}52.9 & \cellcolor{momayellow}76.3 & \cellcolor{momayellow}51.9 & \cellcolor{momayellow}65.4 \\

\midrule

\multirow{4}{*}{\rotatebox{90}{\textbf{Ours}}} 
 & Ours-LLaVA & \cellcolor{momapink}81.3 & \cellcolor{momapink}\underline{88.4} & \cellcolor{momapink}\underline{87.1} & \cellcolor{momapink}95.8 & \cellcolor{momayellow}\underline{73.6} & \cellcolor{momayellow}\underline{81.1} & \cellcolor{momayellow}64.4 & \cellcolor{momayellow}\underline{90.3} \\
 & \quad \textit{\color{gray} w/ Noisy Pose} & \cellcolor{momapink}\color{gray}79.5 & \cellcolor{momapink}\color{gray}87.5 & \cellcolor{momapink}\color{gray}84.4 & \cellcolor{momapink}\color{gray}93.6 & \cellcolor{momayellow}\color{gray}72.2 & \cellcolor{momayellow}\color{gray}80.2 & \cellcolor{momayellow}\color{gray}61.2 & \cellcolor{momayellow}\color{gray}87.1 \\
 & Ours-GPT & \cellcolor{momapink}\textbf{83.0} & \cellcolor{momapink}\textbf{90.2} & \cellcolor{momapink}86.2 & \cellcolor{momapink}\textbf{97.9} & \cellcolor{momayellow}\textbf{75.0} & \cellcolor{momayellow}\textbf{83.0} & \cellcolor{momayellow}\textbf{65.7} & \cellcolor{momayellow}\textbf{91.4} \\
 & \quad \textit{\color{gray} w/ Noisy Pose} & \cellcolor{momapink}\color{gray}82.1 & \cellcolor{momapink}\color{gray}89.3 & \cellcolor{momapink}\color{gray}86.7 & \cellcolor{momapink}\color{gray}95.7 & \cellcolor{momayellow}\color{gray}73.6 & \cellcolor{momayellow}\color{gray}82.1 & \cellcolor{momayellow}\color{gray}62.5 & \cellcolor{momayellow}\color{gray}87.3 \\
\bottomrule
\end{tabular}%
}
\end{table*}

\noindent\textbf{2) Relationship Prediction (Edge).} 
Our method demonstrates superior performance on the challenging predicate prediction task. 
Although our edge generation paradigm resembles ConceptGraphs, our performance leap stems fundamentally from enhanced object node accuracy. 
Since relationship reasoning depends on anchor objects, our precise, less hallucinated node captions provide the LLM with superior context for inferring spatial interactions. 
This effectively prevents cascading errors—where misclassified objects yield nonsensical relationships—that plague passive-observation baselines. 
Consequently, even without a heavy relational reasoning module, \textbf{Ours-GPT} achieves 97.9\% R@10 on SceneFun3D (vs. OpenFunGraph's 96.2\%) and 65.7\% R@5 on FunGraph3D (vs. 65.1\%). 
These gains confirm that resolving object-level semantic ambiguity is critical for improving overall scene graph connectivity.

\noindent\textbf{3) Mechanism Analysis.} 
We analyze the variants in Table~\ref{tab:main_results}  to examine the main components.
\textbf{(a) Node Quality Drives Edge Gains:} \textit{Ours Node + CG Edge} yields relationship predictions close to our full model (95.6\% vs 97.9\% R@10 on SceneFun3D) despite using the vanilla ConceptGraphs edge builder. 
This result indicates that the edge improvements are largely driven by enhanced object-node accuracy.
\textbf{(b) Passive Observation Bottleneck:} \textit{Passive VLM Baseline} (top-3 confident crops from passive tracks) underperforms our method by large margins. This shows that perception bottlenecks from viewpoint constraints cannot be resolved by simply scaling up backend foundation models.
\textbf{(c) Failure of Isolated Re-shots:} Isolated re-shot captioning (\textit{Re-shot VLM}) leads to drastic semantic degradation (dropping to 59.4\% Object R@3 on SceneFun3D). This underscores that the tight synergy of our image concatenation and uncertainty-guided RAG is essential to prevent reshot degradation.

\noindent\textbf{4) Robustness against Geometric Degradation.}
We stress-test our system by injecting camera pose noise during reconstruction to simulate tracking drift. 
As shown in \cref{tab:main_results} (\textit{w/ Noisy Pose}), our framework exhibits exceptional resilience: the Object R@3 of \textbf{Ours-GPT} on SceneFun3D drops by merely 0.9\% (83.0\% to 82.1\%), and \textbf{Ours-LLaVA} degrades by less than $\sim$3.2\% on FunGraph3D. 
This disproves the concern that active re-shooting over-relies on perfect 3D reconstructions. 
This robustness stems from: (1) VLM's inherent perceptual tolerance to geometric ghosting when global semantic completeness is preserved; (2) our image concatenation strategy, which allows the VLM to fall back on the uncorrupted 2D crop as a dependable reference even under severe structural distortion.

\subsection{Fine-grained Semantic Quality: Human Evaluation}
\label{sec:human_eval}

While automated metrics assess recall against predefined ground truths, the open-vocabulary nature of our method inherently complicates automatic evaluation. 
Following ConceptGraphs~\cite{gu2024conceptgraphs}, we resort to a fine-grained human evaluation to rigorously assess the semantic precision of the generated graphs. 
We conduct this qualitative audit on 8 scenes from the \textbf{Replica}~\cite{straub2019replica}, leveraging its high-fidelity geometry as an unambiguous reference, and compare our method against ConceptGraphs (CG) and ConceptGraphs-Detector (CG-D)~\cite{gu2024conceptgraphs}.

\begin{table*}[t]
\centering
\caption{\textbf{Human Evaluation on Replica~\cite{straub2019replica} dataset.} We evaluate fine-grained semantic quality across 8 indoor scenes. \colorbox{aliceblue}{Blue columns} denote Node-level metrics, and \colorbox{honeydew}{Green columns} denote Edge-level metrics. \textbf{Prec.}: Precision; \textbf{Valid}: Count of valid objects; \textbf{Dup.}: Count of duplicate objects. \textbf{Bold} indicates best performance.}
\label{tab:human_eval}

\renewcommand{\arraystretch}{1.2}
\setlength{\tabcolsep}{3.5pt} 

\resizebox{\textwidth}{!}{%
\begin{tabular}{l ccc ccc ccc | ccc ccc}
\toprule
\multirow{3}{*}{\textbf{Scene}} & \multicolumn{9}{c}{\cellcolor{aliceblue}\textbf{Node Quality (Objects)}} & \multicolumn{6}{c}{\cellcolor{honeydew}\textbf{Edge Quality (Relationships)}} \\
\cmidrule(lr){2-10} \cmidrule(lr){11-16}
 & \multicolumn{3}{c}{\cellcolor{aliceblue}Precision $\uparrow$} & \multicolumn{3}{c}{\cellcolor{aliceblue}Valid Count $\uparrow$} & \multicolumn{3}{c}{\cellcolor{aliceblue}Duplicates $\downarrow$} & \multicolumn{3}{c}{\cellcolor{honeydew}Precision $\uparrow$} & \multicolumn{3}{c}{\cellcolor{honeydew}Valid Count $\uparrow$} \\
\cmidrule(lr){2-4} \cmidrule(lr){5-7} \cmidrule(lr){8-10} \cmidrule(lr){11-13} \cmidrule(lr){14-16}
 & \cellcolor{aliceblue}Ours & \cellcolor{aliceblue}CG & \cellcolor{aliceblue}CG-D & \cellcolor{aliceblue}Ours & \cellcolor{aliceblue}CG & \cellcolor{aliceblue}CG-D & \cellcolor{aliceblue}Ours & \cellcolor{aliceblue}CG & \cellcolor{aliceblue}CG-D & \cellcolor{honeydew}Ours & \cellcolor{honeydew}CG & \cellcolor{honeydew}CG-D & \cellcolor{honeydew}Ours & \cellcolor{honeydew}CG & \cellcolor{honeydew}CG-D \\
\midrule

room0 & \cellcolor{aliceblue}\textbf{0.87} & \cellcolor{aliceblue}0.77 & \cellcolor{aliceblue}0.53 & \cellcolor{aliceblue}61 & \cellcolor{aliceblue}57 & \cellcolor{aliceblue}60 & \cellcolor{aliceblue}\textbf{1} & \cellcolor{aliceblue}4 & \cellcolor{aliceblue}3 & \cellcolor{honeydew}\textbf{0.93} & \cellcolor{honeydew}0.87 & \cellcolor{honeydew}0.88 & \cellcolor{honeydew}27 & \cellcolor{honeydew}15 & \cellcolor{honeydew}16 \\
room1 & \cellcolor{aliceblue}\textbf{0.88} & \cellcolor{aliceblue}0.73 & \cellcolor{aliceblue}0.71 & \cellcolor{aliceblue}51 & \cellcolor{aliceblue}45 & \cellcolor{aliceblue}42 & \cellcolor{aliceblue}\textbf{0} & \cellcolor{aliceblue}5 & \cellcolor{aliceblue}3 & \cellcolor{honeydew}\textbf{0.97} & \cellcolor{honeydew}0.92 & \cellcolor{honeydew}0.91 & \cellcolor{honeydew}30 & \cellcolor{honeydew}12 & \cellcolor{honeydew}11 \\
room2 & \cellcolor{aliceblue}\textbf{0.85} & \cellcolor{aliceblue}0.63 & \cellcolor{aliceblue}0.50 & \cellcolor{aliceblue}47 & \cellcolor{aliceblue}48 & \cellcolor{aliceblue}50 & \cellcolor{aliceblue}\textbf{0} & \cellcolor{aliceblue}3 & \cellcolor{aliceblue}2 & \cellcolor{honeydew}\textbf{0.94} & \cellcolor{honeydew}0.91 & \cellcolor{honeydew}0.92 & \cellcolor{honeydew}35 & \cellcolor{honeydew}11 & \cellcolor{honeydew}12 \\
office0 & \cellcolor{aliceblue}\textbf{0.73} & \cellcolor{aliceblue}0.61 & \cellcolor{aliceblue}0.61 & \cellcolor{aliceblue}48 & \cellcolor{aliceblue}44 & \cellcolor{aliceblue}41 & \cellcolor{aliceblue}\textbf{1} & \cellcolor{aliceblue}1 & \cellcolor{aliceblue}1 & \cellcolor{honeydew}\textbf{0.93} & \cellcolor{honeydew}0.78 & \cellcolor{honeydew}0.82 & \cellcolor{honeydew}27 & \cellcolor{honeydew}9 & \cellcolor{honeydew}11 \\
office1 & \cellcolor{aliceblue}\textbf{0.73} & \cellcolor{aliceblue}0.64 & \cellcolor{aliceblue}0.46 & \cellcolor{aliceblue}44 & \cellcolor{aliceblue}25 & \cellcolor{aliceblue}24 & \cellcolor{aliceblue}\textbf{0} & \cellcolor{aliceblue}1 & \cellcolor{aliceblue}3 & \cellcolor{honeydew}\textbf{0.93} & \cellcolor{honeydew}0.80 & \cellcolor{honeydew}0.86 & \cellcolor{honeydew}28 & \cellcolor{honeydew}5 & \cellcolor{honeydew}7 \\
office2 & \cellcolor{aliceblue}\textbf{0.87} & \cellcolor{aliceblue}0.77 & \cellcolor{aliceblue}0.68 & \cellcolor{aliceblue}67 & \cellcolor{aliceblue}48 & \cellcolor{aliceblue}44 & \cellcolor{aliceblue}\textbf{1} & \cellcolor{aliceblue}3 & \cellcolor{aliceblue}2 & \cellcolor{honeydew}\textbf{0.88} & \cellcolor{honeydew}0.79 & \cellcolor{honeydew}0.86 & \cellcolor{honeydew}34 & \cellcolor{honeydew}14 & \cellcolor{honeydew}14 \\
office3 & \cellcolor{aliceblue}\textbf{0.85} & \cellcolor{aliceblue}0.69 & \cellcolor{aliceblue}0.60 & \cellcolor{aliceblue}65 & \cellcolor{aliceblue}59 & \cellcolor{aliceblue}57 & \cellcolor{aliceblue}2 & \cellcolor{aliceblue}4 & \cellcolor{aliceblue}2 & \cellcolor{honeydew}\textbf{0.84} & \cellcolor{honeydew}0.78 & \cellcolor{honeydew}0.77 & \cellcolor{honeydew}32 & \cellcolor{honeydew}9 & \cellcolor{honeydew}13 \\
office4 & \cellcolor{aliceblue}\textbf{0.79} & \cellcolor{aliceblue}0.61 & \cellcolor{aliceblue}0.57 & \cellcolor{aliceblue}53 & \cellcolor{aliceblue}41 & \cellcolor{aliceblue}46 & \cellcolor{aliceblue}\textbf{1} & \cellcolor{aliceblue}5 & \cellcolor{aliceblue}4 & \cellcolor{honeydew}\textbf{0.86} & \cellcolor{honeydew}0.67 & \cellcolor{honeydew}0.80 & \cellcolor{honeydew}22 & \cellcolor{honeydew}3 & \cellcolor{honeydew}5 \\

\midrule
\rowcolor{gray!20} 
Average & \textbf{0.82} & 0.68 & 0.58 & - & - & - & - & - & - & \textbf{0.91} & 0.82 & 0.85 & - & - & - \\
\bottomrule
\end{tabular}%
}
\end{table*}

\noindent\textbf{Strict Annotation Protocol.}
To address the ambiguities inherent in open-vocabulary generation, we established a strict annotation protocol. 
We recruited three independent domain experts (students in 3D vision) to evaluate the randomized base units (point clouds, images, and predicted captions). 
Crucially, we defined a clear boundary for ``accuracy'': a node or edge caption is deemed \textit{Valid} if it captures the correct semantic category and primary spatial/functional attributes. 
Acceptable paraphrases and synonymous descriptions (e.g., ``a wooden table'' vs. ``a brown desk'') are treated as equally valid semantic equivalents, provided they do not introduce hallucinated attributes or incorrect spatial logic.

\noindent\textbf{Inter-Annotator Agreement (IAA).}
To ensure the reliability of the human labels, we conducted a consistency check among the three annotators. 
We computed the Fleiss' Kappa ($\kappa$) score across a subset of 100 randomly sampled evaluation units. 
The annotators achieved a $\kappa$ score of 0.76, indicating ``substantial agreement.'' 
Disagreements primarily stemmed from ambiguous linguistic expressions or subjective interpretations of spatial relationships, which were subsequently resolved via majority voting for the final tally.

\noindent\textbf{Results Analysis.}
\Cref{tab:human_eval} reports the comprehensive evaluation across node and edge qualities.
Our method consistently outperforms both ConceptGraphs (CG) and ConceptGraphsDetector (CG-D) across most evaluation metrics. In terms of node precision, our method achieves an
average score of 0.82, which is notably higher than CG (0.68) and CG-D (0.58), demonstrating the
effectiveness of our re-shot guided uncertainty estimation in reducing noise during object caption aggregation. For edge precision, our method also achieves the highest average score (0.91), surpassing
CG (0.82) and CG-D (0.85), indicating that our structured prompt and refined relationship categories lead to more accurate and interpretable edge captions. In addition, our method substantially reduces duplicate predictions while maintaining a higher number of valid objects and edges, further confirming its robustness.

\begin{figure}[ht]
\begin{center}
\includegraphics[width=\linewidth]{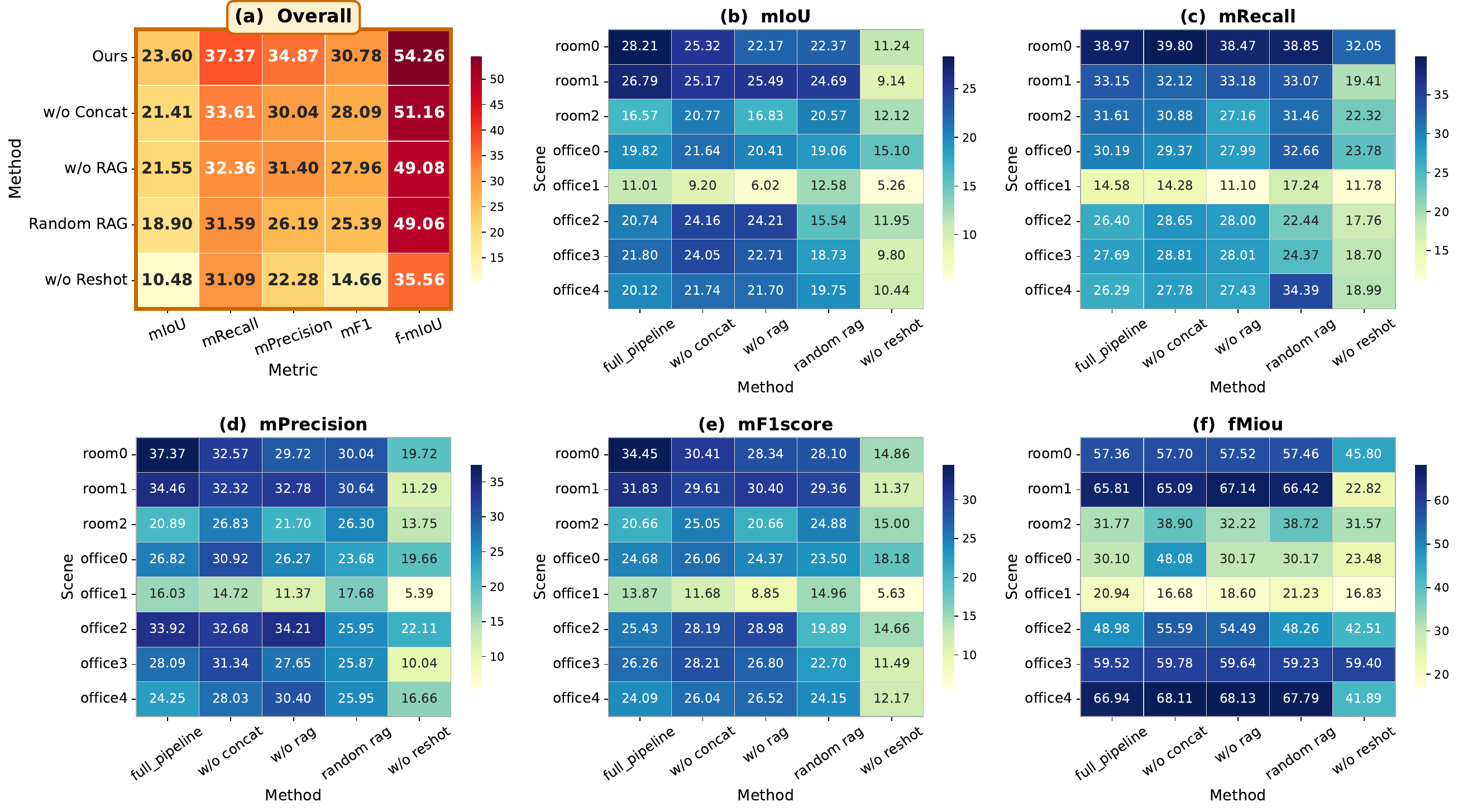}
\end{center}
\caption{
Ablation study quantifying component contributions. Performance degradation is observed when removing reshot guided uncertainty estimation (w/o Reshot),
RAG (w/o RAG), concatenated re-shot images
(w/o Concat), or using random retrieval (Random
RAG), confirming the complementary roles of all
proposed components.
}
\label{fig:ablation}
\end{figure}

\subsection{Ablation Study}
We conduct ablation studies to quantify the contribution of each component in our pipeline. 
Our full model (\textit{Ours}) consists of dynamic downsample-mapping \& fusion, re-shot guided uncertainty estimation, and node-level RAG with concatenated re-shot image prompts. 
We compare against several variants: (i) removing re-shot guided uncertainty estimation (\textit{w/o Reshot}), (ii) removing RAG (\textit{w/o RAG}), (iii) applying random retrieval for RAG (\textit{random-RAG}), and (iv) removing the concatenated re-shot image prompts (\textit{w/o Concat}). 

Experiments are conducted on the Replica dataset \cite{straub2019replica}. 
Following the protocol used in \cite{gu2024conceptgraphs, jatavallabhula2023conceptfusion}, we obtain per-point semantic GT labels by fusing 2D masks into 3D via gradSLAM \cite{gradslam}. 
To ensure faithful open-vocabulary evaluation, we use GPT-4o as a semantic assigner to match GT labels with predicted node captions. 
After 1-NN point-cloud matching and confusion matrix construction, quantitative results are reported in \cref{fig:ablation}, where (a) shows overall performance and (b)-(f) detail metric-specific heatmaps across individual scenes. 
Ablation results confirm that removing Reshot, RAG, Concat, or using Random RAG leads to performance drops, highlighting their complementary roles.

As shown in \cref{fig:ablation} (a), removing any component leads to a performance drop, confirming their complementary roles. 
Notably, removing re-shot guided uncertainty estimation (\textit{w/o Reshot}) causes the most drastic degradation, e.g., in mF1 (14.66 vs. 30.78) and f-mIoU (35.56 vs. 54.26). 
This is because \textit{Reshot} acts as a fundamental filter; without it, the system is forced to fuse unreliable or hallucinated captions, which severely degrades the global semantic map.
Eliminating RAG (\textit{w/o RAG}) reduces both precision and overall accuracy, while replacing it with random retrieval (\textit{random-RAG}) further deteriorates performance, highlighting the necessity of reliable retrieval. 
Removing the concatenated re-shot image (\textit{w/o Concat}) also leads to lower recall and f-mIoU, suggesting that multi-view prompts alleviate viewpoint bias and enrich object descriptions. 
These results collectively demonstrate that all three proposed components contribute significantly to the robustness and accuracy of our framework.

\subsection{Real-World Robotic Deployment}
\label{sec:real_world}

To further validate the practical viability of our approach, we deploy \ours \space on a mobile robot and conduct experiments in a real-world indoor facility.

\begin{figure}[ht]
\begin{center}
\includegraphics[width=\linewidth]{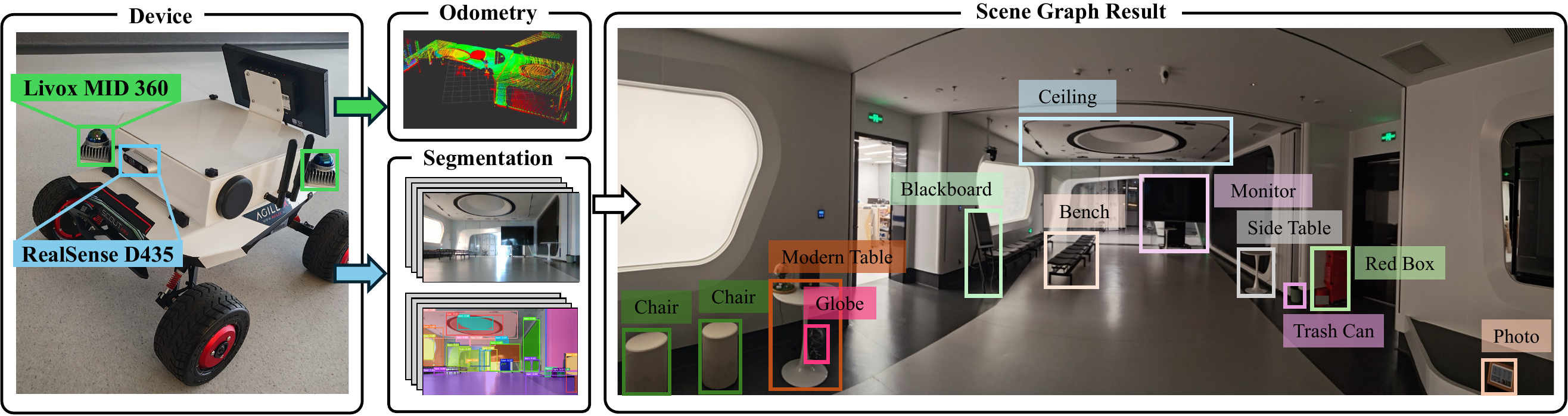}
\end{center}
\caption{
\textbf{Real-world robotic deployment.} 
By fusing RGB-D perception with LiDAR-IMU odometry, our mobile platform successfully drives the \ours~framework to generate high-fidelity semantic 3D scene graphs in real-world indoor environments.
}
\label{fig:realworld}
\end{figure}

\noindent\textbf{Hardware Setup.} 
As illustrated in \cref{fig:realworld}, our custom robotic platform is built upon an Agilex SCOUT MINI chassis. 
The sensor suite comprises a RealSense D435 RGB-D camera for visual perception, a CH110 IMU, and dual Livox MID 360 LiDARs to provide omnidirectional geometric coverage and blind-spot reduction. 
All data processing and sensor synchronization are handled by an onboard NVIDIA Jetson AGX Xavier computer.

\noindent\textbf{Deployment Pipeline.} 
During the physical deployment, the robot teleoperates through the facility, continuously capturing RGB and Depth streams. 
Simultaneously, we maintain a LiDAR-IMU Odometry (LIO) pipeline to estimate real-time camera poses. 
These poses are strictly utilized to perform cross-frame matching and tracking of 2D segmented object instances. 
Once the data collection is complete, the fused 3D object observations are fed into our RAG-3DSG framework to generate the final textual 3D semantic scene graph.

\noindent\textbf{Results and Robustness Analysis.} 
Aligning with our stress tests, physical deployment reveals the system's profound resilience. 
Under inferior localization (e.g., drift-prone pure IMU), corrupted point clouds degrade the active \textit{re-shot} mechanism via ghosting. 
Yet, the system only degrades marginally, leveraging the RAG module's semantic anchors for robust contextual refinement. 
Conversely, integrating robust state-estimation (e.g., Fast-LIO2 \cite{xu2022fast}) unleashes the full synergy of our re-shot and RAG modules, yielding high-fidelity scene graphs.

\subsection{Runtime Analysis and System Scalability}
\label{sec:runtime_analysis}
To address the computational overhead nature of foundation models, we strictly optimize our framework for practical scalability. 
While currently operating as an offline system, our dynamic downsample-mapping strategy significantly accelerates the initial 3D fusion phase by pruning pointcloud, reducing the per-iteration processing time by nearly two-thirds compared to ConceptGraphs~\cite{gu2024conceptgraphs} (e.g., from 6.65s to 2.49s on Replica). 
For semantic reasoning, the computational and monetary costs remain highly manageable. 
Our LLaVA variant runs efficiently on a single local consumer GPU, while the GPT-4o variant incurs a minimal API cost of $\sim$\$0.50 per scene on FunGraph3D. 
Under the same backbone as ConceptGraphs (VLM reasoning costs of $\sim$0.78s/iteration), the additional active verification introduces a rendering cost of $\sim$0.26s/object. 
Finally, although designed for offline high-precision generation, our pipeline aligns well with modern decoupled robotic architectures (e.g., DAAAM~\cite{gorlo2025describe}). This paradigm allows our method to be seamlessly integrated as a low-frequency semantic backend to guide high-frequency real-time exploration in future online applications.
\section{CONCLUSION}
In this work, we present \ours \space for generating accurate and robust open-vocabulary 3D scene graphs. 
We are the first to address semantic inconsistency in cross-view aggregation by introducing active re-shot verification and an object-level RAG module for contextual refinement. 
Extensive evaluations on real-world datasets demonstrate new state-of-the-art performance in both node and edge accuracy, while human evaluation confirms the generation of cleaner, less hallucinated graphs. 
Crucially, extreme geometric stress tests and physical robotic deployment validate our system's exceptional resilience against sensor noise, paving the way for reliable 3D perception in complex downstream tasks.

%
\section*{Acknowledgements}
Sihong Xie was supported by the National Key R\&D Program of China (Grant No.\ 2023YFF0725001), the Department of Science and Technology of Guangdong Province (2023CX10X079), the Guangzhou-HKUST (GZ) Joint Funding Program (Grant No.\ 2023A03J0008), and the Education Bureau Guangzhou Municipality, to which the authors extend their sincere gratitude for the support.

%
%
\bibliographystyle{splncs04}
\bibliography{main}

@String(ECCV  = {Eur. Conf. Comput. Vis.})

@article{li2024scene,
  title={Scene graph generation: A comprehensive survey},
  author={Li, Hongsheng and Zhu, Guangming and Zhang, Liang and Jiang, Youliang and Dang, Yixuan and Hou, Haoran and Shen, Peiyi and Zhao, Xia and Shah, Syed Afaq Ali and Bennamoun, Mohammed},
  journal={Neurocomputing},
  volume={566},
  pages={127052},
  year={2024},
  publisher={Elsevier}
}

@inproceedings{sun2023evidential,
  title={Evidential uncertainty and diversity guided active learning for scene graph generation},
  author={Sun, Shuzhou and Zhi, Shuaifeng and Heikkil{\"a}, Janne and Liu, Li},
  booktitle={The Eleventh International Conference on Learning Representations},
  year={2023}
}

@inproceedings{liu2021fully,
  title={Fully convolutional scene graph generation},
  author={Liu, Hengyue and Yan, Ning and Mortazavi, Masood and Bhanu, Bir},
  booktitle={Proceedings of the IEEE/CVF conference on computer vision and pattern recognition},
  pages={11546--11556},
  year={2021}
}

@inproceedings{yin2018zoom,
  title={Zoom-net: Mining deep feature interactions for visual relationship recognition},
  author={Yin, Guojun and Sheng, Lu and Liu, Bin and Yu, Nenghai and Wang, Xiaogang and Shao, Jing and Loy, Chen Change},
  booktitle={Proceedings of the European conference on computer vision (ECCV)},
  pages={322--338},
  year={2018}
}

@article{krishna2017visual,
  title={Visual genome: Connecting language and vision using crowdsourced dense image annotations},
  author={Krishna, Ranjay and Zhu, Yuke and Groth, Oliver and Johnson, Justin and Hata, Kenji and Kravitz, Joshua and Chen, Stephanie and Kalantidis, Yannis and Li, Li-Jia and Shamma, David A and others},
  journal={International journal of computer vision},
  volume={123},
  number={1},
  pages={32--73},
  year={2017},
  publisher={Springer}
}

@inproceedings{lu2016visual,
  title={Visual relationship detection with language priors},
  author={Lu, Cewu and Krishna, Ranjay and Bernstein, Michael and Fei-Fei, Li},
  booktitle={European conference on computer vision},
  pages={852--869},
  year={2016},
  organization={Springer}
}

@inproceedings{johnson2015image,
  title={Image retrieval using scene graphs},
  author={Johnson, Justin and Krishna, Ranjay and Stark, Michael and Li, Li-Jia and Shamma, David and Bernstein, Michael and Fei-Fei, Li},
  booktitle={Proceedings of the IEEE conference on computer vision and pattern recognition},
  pages={3668--3678},
  year={2015}
}

@inproceedings{agia2022taskography,
  title={Taskography: Evaluating robot task planning over large 3d scene graphs},
  author={Agia, Christopher and Jatavallabhula, Krishna Murthy and Khodeir, Mohamed and Miksik, Ondrej and Vineet, Vibhav and Mukadam, Mustafa and Paull, Liam and Shkurti, Florian},
  booktitle={Conference on Robot Learning},
  pages={46--58},
  year={2022},
  organization={PMLR}
}

@article{rana2023sayplan,
  title={Sayplan: Grounding large language models using 3d scene graphs for scalable robot task planning},
  author={Rana, Krishan and Haviland, Jesse and Garg, Sourav and Abou-Chakra, Jad and Reid, Ian and Suenderhauf, Niko},
  journal={arXiv preprint arXiv:2307.06135},
  year={2023}
}

@article{rosinol2021kimera,
  title={Kimera: From SLAM to spatial perception with 3D dynamic scene graphs},
  author={Rosinol, Antoni and Violette, Andrew and Abate, Marcus and Hughes, Nathan and Chang, Yun and Shi, Jingnan and Gupta, Arjun and Carlone, Luca},
  journal={The International Journal of Robotics Research},
  volume={40},
  number={12-14},
  pages={1510--1546},
  year={2021},
  publisher={SAGE Publications Sage UK: London, England}
}

@article{hughes2022hydra,
  title={Hydra: A real-time spatial perception system for 3D scene graph construction and optimization},
  author={Hughes, Nathan and Chang, Yun and Carlone, Luca},
  journal={arXiv preprint arXiv:2201.13360},
  year={2022}
}

@inproceedings{wu2021scenegraphfusion,
  title={Scenegraphfusion: Incremental 3d scene graph prediction from rgb-d sequences},
  author={Wu, Shun-Cheng and Wald, Johanna and Tateno, Keisuke and Navab, Nassir and Tombari, Federico},
  booktitle={Proceedings of the IEEE/CVF Conference on Computer Vision and Pattern Recognition},
  pages={7515--7525},
  year={2021}
}

@inproceedings{armeni20193d,
  title={3d scene graph: A structure for unified semantics, 3d space, and camera},
  author={Armeni, Iro and He, Zhi-Yang and Gwak, JunYoung and Zamir, Amir R and Fischer, Martin and Malik, Jitendra and Savarese, Silvio},
  booktitle={Proceedings of the IEEE/CVF international conference on computer vision},
  pages={5664--5673},
  year={2019}
}

@article{kim20193,
  title={3-D scene graph: A sparse and semantic representation of physical environments for intelligent agents},
  author={Kim, Ue-Hwan and Park, Jin-Man and Song, Taek-Jin and Kim, Jong-Hwan},
  journal={IEEE transactions on cybernetics},
  volume={50},
  number={12},
  pages={4921--4933},
  year={2019},
  publisher={IEEE}
}

@inproceedings{wald2020learning,
  title={Learning 3d semantic scene graphs from 3d indoor reconstructions},
  author={Wald, Johanna and Dhamo, Helisa and Navab, Nassir and Tombari, Federico},
  booktitle={Proceedings of the IEEE/CVF Conference on Computer Vision and Pattern Recognition},
  pages={3961--3970},
  year={2020}
}

@inproceedings{gu2024conceptgraphs,
  title={Conceptgraphs: Open-vocabulary 3d scene graphs for perception and planning},
  author={Gu, Qiao and Kuwajerwala, Ali and Morin, Sacha and Jatavallabhula, Krishna Murthy and Sen, Bipasha and Agarwal, Aditya and Rivera, Corban and Paul, William and Ellis, Kirsty and Chellappa, Rama and others},
  booktitle={2024 IEEE International Conference on Robotics and Automation (ICRA)},
  pages={5021--5028},
  year={2024},
  organization={IEEE}
}

@inproceedings{werby2024hierarchical,
  title={Hierarchical open-vocabulary 3d scene graphs for language-grounded robot navigation},
  author={Werby, Abdelrhman and Huang, Chenguang and B{\"u}chner, Martin and Valada, Abhinav and Burgard, Wolfram},
  booktitle={First Workshop on Vision-Language Models for Navigation and Manipulation at ICRA 2024},
  year={2024}
}

@inproceedings{koch2024open3dsg,
  title={Open3dsg: Open-vocabulary 3d scene graphs from point clouds with queryable objects and open-set relationships},
  author={Koch, Sebastian and Vaskevicius, Narunas and Colosi, Mirco and Hermosilla, Pedro and Ropinski, Timo},
  booktitle={Proceedings of the IEEE/CVF Conference on Computer Vision and Pattern Recognition},
  pages={14183--14193},
  year={2024}
}

@article{maggio2024clio,
  title={Clio: Real-time task-driven open-set 3d scene graphs},
  author={Maggio, Dominic and Chang, Yun and Hughes, Nathan and Trang, Matthew and Griffith, Dan and Dougherty, Carlyn and Cristofalo, Eric and Schmid, Lukas and Carlone, Luca},
  journal={IEEE Robotics and Automation Letters},
  year={2024},
  publisher={IEEE}
}

@article{jatavallabhula2023conceptfusion,
  title={Conceptfusion: Open-set multimodal 3d mapping},
  author={Jatavallabhula, Krishna Murthy and Kuwajerwala, Alihusein and Gu, Qiao and Omama, Mohd and Chen, Tao and Maalouf, Alaa and Li, Shuang and Iyer, Ganesh and Saryazdi, Soroush and Keetha, Nikhil and others},
  journal={arXiv preprint arXiv:2302.07241},
  year={2023}
}

@inproceedings{kirillov2023segment,
  title={Segment anything},
  author={Kirillov, Alexander and Mintun, Eric and Ravi, Nikhila and Mao, Hanzi and Rolland, Chloe and Gustafson, Laura and Xiao, Tete and Whitehead, Spencer and Berg, Alexander C and Lo, Wan-Yen and others},
  booktitle={Proceedings of the IEEE/CVF international conference on computer vision},
  pages={4015--4026},
  year={2023}
}

@article{hurst2024gpt,
  title={Gpt-4o system card},
  author={Hurst, Aaron and Lerer, Adam and Goucher, Adam P and Perelman, Adam and Ramesh, Aditya and Clark, Aidan and Ostrow, AJ and Welihinda, Akila and Hayes, Alan and Radford, Alec and others},
  journal={arXiv preprint arXiv:2410.21276},
  year={2024}
}

@inproceedings{shridhar2022cliport,
  title={Cliport: What and where pathways for robotic manipulation},
  author={Shridhar, Mohit and Manuelli, Lucas and Fox, Dieter},
  booktitle={Conference on robot learning},
  pages={894--906},
  year={2022},
  organization={PMLR}
}

@inproceedings{rashid2023language,
  title={Language embedded radiance fields for zero-shot task-oriented grasping},
  author={Rashid, Adam and Sharma, Satvik and Kim, Chung Min and Kerr, Justin and Chen, Lawrence Yunliang and Kanazawa, Angjoo and Goldberg, Ken},
  booktitle={7th Annual Conference on Robot Learning},
  year={2023}
}

@article{gadre2022clip,
  title={Clip on wheels: Zero-shot object navigation as object localization and exploration},
  author={Gadre, Samir Yitzhak and Wortsman, Mitchell and Ilharco, Gabriel and Schmidt, Ludwig and Song, Shuran},
  journal={arXiv preprint arXiv:2203.10421},
  volume={3},
  number={4},
  pages={7},
  year={2022}
}

@inproceedings{shah2023lm,
  title={Lm-nav: Robotic navigation with large pre-trained models of language, vision, and action},
  author={Shah, Dhruv and Osi{\'n}ski, B{\l}a{\.z}ej and Levine, Sergey and others},
  booktitle={Conference on robot learning},
  pages={492--504},
  year={2023},
  organization={PMLR}
}

@article{chen2026psg,
  title={PSG-Nav: Probabilistic Scene Graph Navigation via Multiverse Decision Making},
  author={Chen, Rufeng and Chang, Yue and Tang, Xiaqiang and Chen, Hechang and Xie, Sihong},
  journal={arXiv preprint arXiv:2606.01313},
  year={2026}
}

@inproceedings{gay2018visual,
  title={Visual graphs from motion (vgfm): Scene understanding with object geometry reasoning},
  author={Gay, Paul and Stuart, James and Del Bue, Alessio},
  booktitle={Asian Conference on Computer Vision},
  pages={330--346},
  year={2018},
  organization={Springer}
}

@article{straub2019replica,
  title={The replica dataset: A digital replica of indoor spaces},
  author={Straub, Julian and Whelan, Thomas and Ma, Lingni and Chen, Yufan and Wijmans, Erik and Green, Simon and Engel, Jakob J and Mur-Artal, Raul and Ren, Carl and Verma, Shobhit and others},
  journal={arXiv preprint arXiv:1906.05797},
  year={2019}
}

@article{gradslam,
  title={gradSLAM: Automagically differentiable SLAM},
  author={Jatavallabhula, Krishna Murthy and Saryazdi, Soroush and Iyer, Ganesh and Paull, Liam},
  journal={arXiv preprint arXiv:1910.10672},
  year={2019}
}

@inproceedings{vazquez2001viewpoint,
  title={Viewpoint selection using viewpoint entropy.},
  author={V{\'a}zquez, Pere-Pau and Feixas, Miquel and Sbert, Mateu and Heidrich, Wolfgang},
  booktitle={VMV},
  volume={1},
  pages={273--280},
  year={2001}
}

@article{liu2023llava,
  title={Visual instruction tuning},
  author={Liu, Haotian and Li, Chunyuan and Wu, Qingyang and Lee, Yong Jae},
  journal={Advances in neural information processing systems},
  volume={36},
  pages={34892--34916},
  year={2023}
}

@inproceedings{scenefun3d,
  title={Scenefun3d: Fine-grained functionality and affordance understanding in 3d scenes},
  author={Delitzas, Alexandros and Takmaz, Ayca and Tombari, Federico and Sumner, Robert and Pollefeys, Marc and Engelmann, Francis},
  booktitle={Proceedings of the IEEE/CVF Conference on Computer Vision and Pattern Recognition},
  pages={14531--14542},
  year={2024}
}

@inproceedings{fungraph3d,
  title={Open-vocabulary functional 3d scene graphs for real-world indoor spaces},
  author={Zhang, Chenyangguang and Delitzas, Alexandros and Wang, Fangjinhua and Zhang, Ruida and Ji, Xiangyang and Pollefeys, Marc and Engelmann, Francis},
  booktitle={Proceedings of the Computer Vision and Pattern Recognition Conference},
  pages={19401--19413},
  year={2025}
}

@article{honerkamp2024language,
  title={Language-grounded dynamic scene graphs for interactive object search with mobile manipulation},
  author={Honerkamp, Daniel and B{\"u}chner, Martin and Despinoy, Fabien and Welschehold, Tim and Valada, Abhinav},
  journal={IEEE Robotics and Automation Letters},
  year={2024},
  publisher={IEEE}
}

@article{yin2024sg,
  title={Sg-nav: Online 3d scene graph prompting for llm-based zero-shot object navigation},
  author={Yin, Hang and Xu, Xiuwei and Wu, Zhenyu and Zhou, Jie and Lu, Jiwen},
  journal={Advances in neural information processing systems},
  volume={37},
  pages={5285--5307},
  year={2024}
}

@article{zhao2026cov,
  title={CoV: Chain-of-View Prompting for Spatial Reasoning},
  author={Zhao, Haoyu and Liu, Akide and Zhang, Zeyu and Wang, Weijie and Chen, Feng and Zhu, Ruihan and Haffari, Gholamreza and Zhuang, Bohan},
  journal={arXiv preprint arXiv:2601.05172},
  year={2026}
}

@article{zhang2026think3d,
  title={Think3D: Thinking with Space for Spatial Reasoning},
  author={Zhang, Zaibin and Wu, Yuhan and Jia, Lianjie and Wang, Yifan and Zhang, Zhongbo and Li, Yijiang and Ran, Binghao and Zhang, Fuxi and Sun, Zhuohan and Yin, Zhenfei and others},
  journal={arXiv preprint arXiv:2601.13029},
  year={2026}
}

@inproceedings{linok2025beyond,
  title={Beyond bare queries: Open-vocabulary object grounding with 3d scene graph},
  author={Linok, Sergey and Zemskova, Tatiana and Ladanova, Svetlana and Titkov, Roman and Yudin, Dmitry and Monastyrny, Maxim and Valenkov, Aleksei},
  booktitle={2025 IEEE International Conference on Robotics and Automation (ICRA)},
  pages={13582--13589},
  year={2025},
  organization={IEEE}
}

@article{yan2025dynamic,
  title={Dynamic open-vocabulary 3d scene graphs for long-term language-guided mobile manipulation},
  author={Yan, Zhijie and Li, Shufei and Wang, Zuoxu and Wu, Lixiu and Wang, Han and Zhu, Jun and Chen, Lijiang and Liu, Jihong},
  journal={IEEE Robotics and Automation Letters},
  year={2025},
  publisher={IEEE}
}

@article{ju2025momagraph,
  title={MomaGraph: State-Aware Unified Scene Graphs with Vision-Language Model for Embodied Task Planning},
  author={Ju, Yuanchen and Liang, Yongyuan and Wang, Yen-Jen and Gireesh, Nandiraju and Ju, Yuanliang and Lee, Seungjae and Gu, Qiao and Hsieh, Elvis and Huang, Furong and Sreenath, Koushil},
  journal={arXiv preprint arXiv:2512.16909},
  year={2025}
}

@inproceedings{yamazaki2024open,
  title={Open-fusion: Real-time open-vocabulary 3d mapping and queryable scene representation},
  author={Yamazaki, Kashu and Hanyu, Taisei and Vo, Khoa and Pham, Thang and Tran, Minh and Doretto, Gianfranco and Nguyen, Anh and Le, Ngan},
  booktitle={2024 IEEE International Conference on Robotics and Automation (ICRA)},
  pages={9411--9417},
  year={2024},
  organization={IEEE}
}

@article{zou2023segment,
  title={Segment everything everywhere all at once},
  author={Zou, Xueyan and Yang, Jianwei and Zhang, Hao and Li, Feng and Li, Linjie and Wang, Jianfeng and Wang, Lijuan and Gao, Jianfeng and Lee, Yong Jae},
  journal={Advances in neural information processing systems},
  volume={36},
  pages={19769--19782},
  year={2023}
}

@inproceedings{radford2021learning,
  title={Learning transferable visual models from natural language supervision},
  author={Radford, Alec and Kim, Jong Wook and Hallacy, Chris and Ramesh, Aditya and Goh, Gabriel and Agarwal, Sandhini and Sastry, Girish and Askell, Amanda and Mishkin, Pamela and Clark, Jack and others},
  booktitle={International conference on machine learning},
  pages={8748--8763},
  year={2021},
  organization={PmLR}
}

@article{gorlo2025describe,
  title={Describe Anything Anywhere At Any Moment},
  author={Gorlo, Nicolas and Schmid, Lukas and Carlone, Luca},
  journal={arXiv preprint arXiv:2512.00565},
  year={2025}
}

@article{xu2022fast,
  title={Fast-lio2: Fast direct lidar-inertial odometry},
  author={Xu, Wei and Cai, Yixi and He, Dongjiao and Lin, Jiarong and Zhang, Fu},
  journal={IEEE Transactions on Robotics},
  volume={38},
  number={4},
  pages={2053--2073},
  year={2022},
  publisher={IEEE}
}

\clearpage
\appendix

\section{Supplementary Material}
\label{Supplementary Material}
In this Supplementary Material, we present additional methodological details, extensive results, and comprehensive system analyses to support the claims made in the main paper. The contents are organized as follows.

\subsection*{Organization}
\begin{itemize}
    \item \textbf{\Cref{appendix:Re-shot Images}}: Examples of re-shot images selected by our method.
    \item \textbf{\Cref{subsec:Case Study}}: Concrete examples of re-shot guided uncertainty estimation.
    \item \textbf{\Cref{supp:hyperparameter}}: Detailed hyperparameter ablation studies.
    \item \textbf{\Cref{supp:real_world_details}}: Extended details for real-world robotic deployment.
    \item \textbf{\Cref{Dynamic Downsample}}: Intuitive examples of the dynamic downsampling strategy.
    \item \textbf{\Cref{appendix:Pseudo Code}}: Detailed pseudo code for the Cross-image Object Mapping.
    \item \textbf{\Cref{appendix:Prompt}}: Complete prompt templates utilized for reasoning.
\end{itemize}

\subsection{Re-shot Images}
\label{appendix:Re-shot Images}
\begin{figure}[ht]
\begin{center}
\includegraphics[width=1\linewidth]{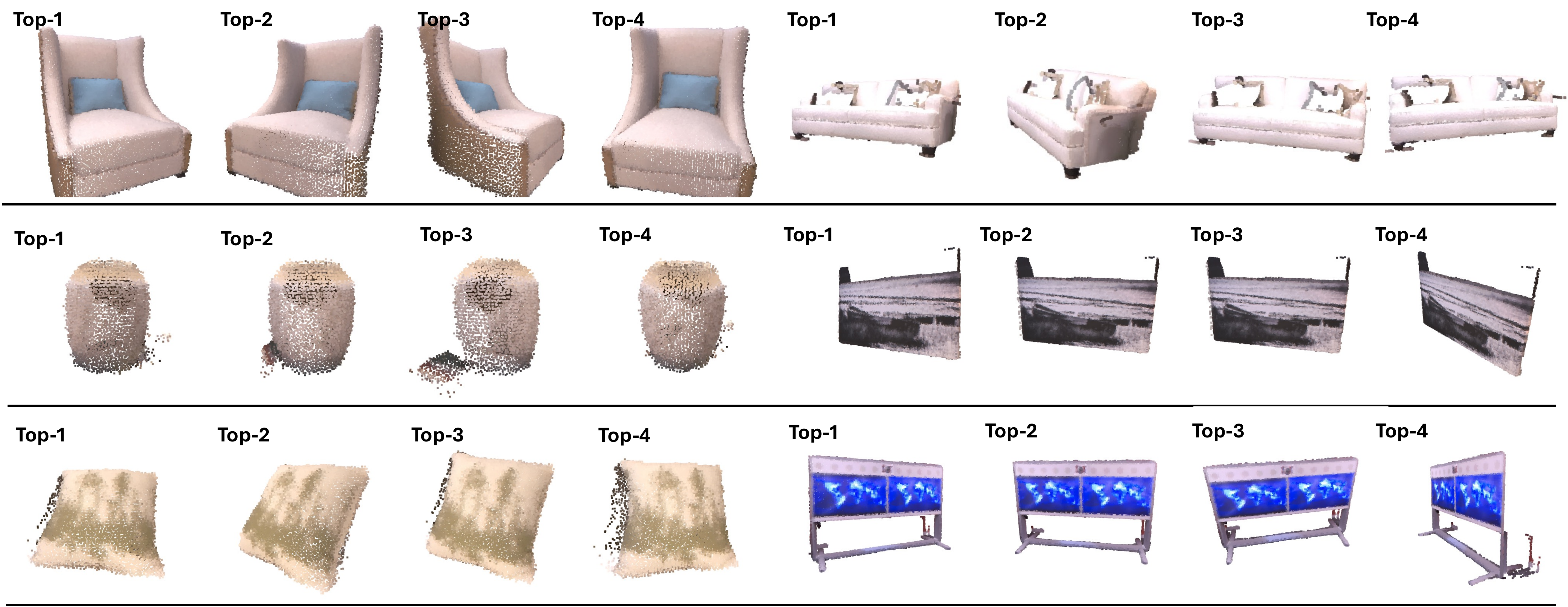}
\end{center}
\caption{
Examples of re-shot images automatically selected by our method. For each object category, we present the top-4 rendered views ranked by our re-shot scoring strategy. From top to bottom, the rows display: an armchair and a sofa (first row), a vase and an artwork (second row), and a cushion and a TV stand (third row).
}
\label{fig:reshot_images}
\end{figure}

Figure~\ref{fig:reshot_images} shows some examples of re-shot images automatically selected by our method.
The initial image crops of an object may suffer from occlusions or constrained viewpoints, 
leading to unreliable captions and noisy semantics. 
To address this issue, we propose a re-shot strategy that leverages object-level point clouds to render new views from arbitrary perspectives, 
ensuring that the essential appearance of the object are faithfully captured. 
This approach eliminates occlusion and viewpoint constraints inherent in the original images.

\subsection{Case Study}
\label{subsec:Case Study}
\begin{figure}[ht]
\begin{center}
\includegraphics[width=1\linewidth]{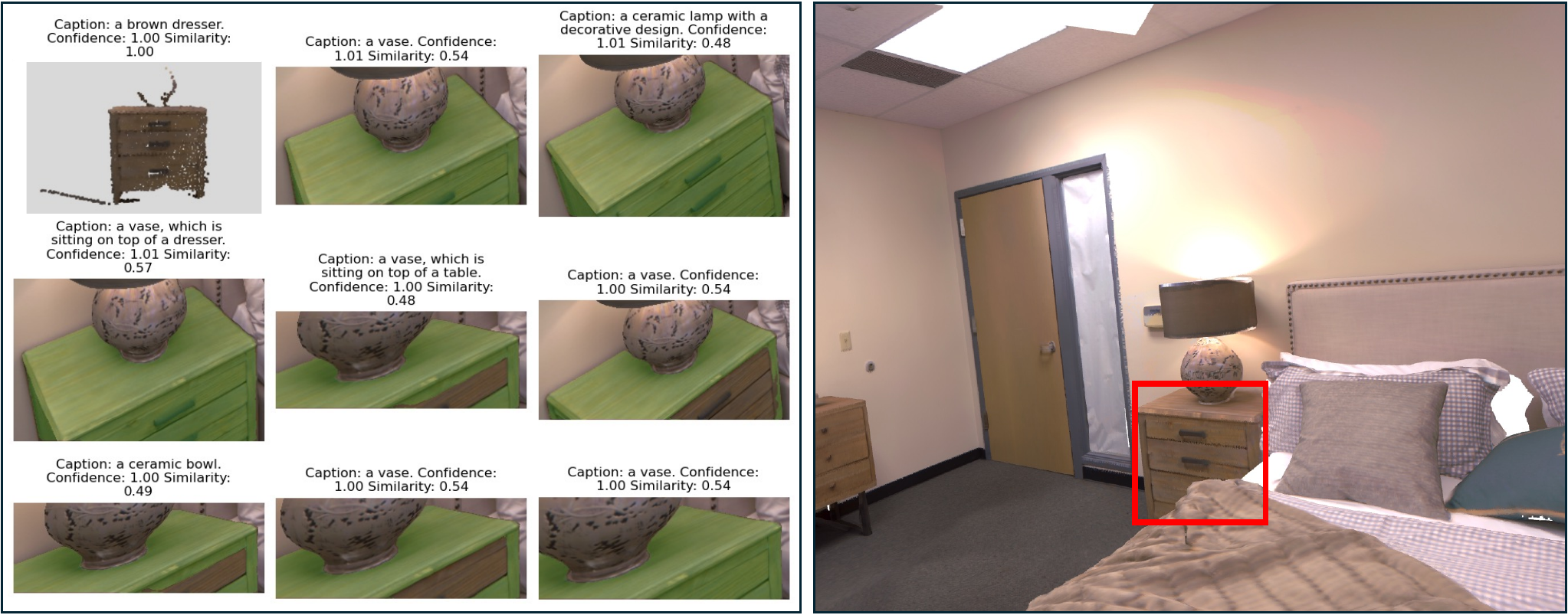}
\end{center}
\caption{
Case study of re-shot guided uncertainty estimation using LLaVA~\cite{liu2023llava} as VLM. 
}
\label{fig:case}
\end{figure}

Figure~\ref{fig:case} demonstrates a concrete example of our re-shot guided uncertainty estimation when LLaVA~\cite{liu2023llava} is the VLM. 
Multiple crop images incorrectly identify the object as \textit{vase} due to occlusions and constrained viewing angles.
In contrast, the re-shot image (top-left) provides a correct caption from an optimal viewpoint selected from our algorithm, avoiding the viewpoint limitations and occlusions that plague the original crop images.
The low similarity scores flag these crop captions as unreliable, triggering our object-level RAG caption refinement. 

\begin{figure}[ht]
  \centering
  \includegraphics[width=1\linewidth]{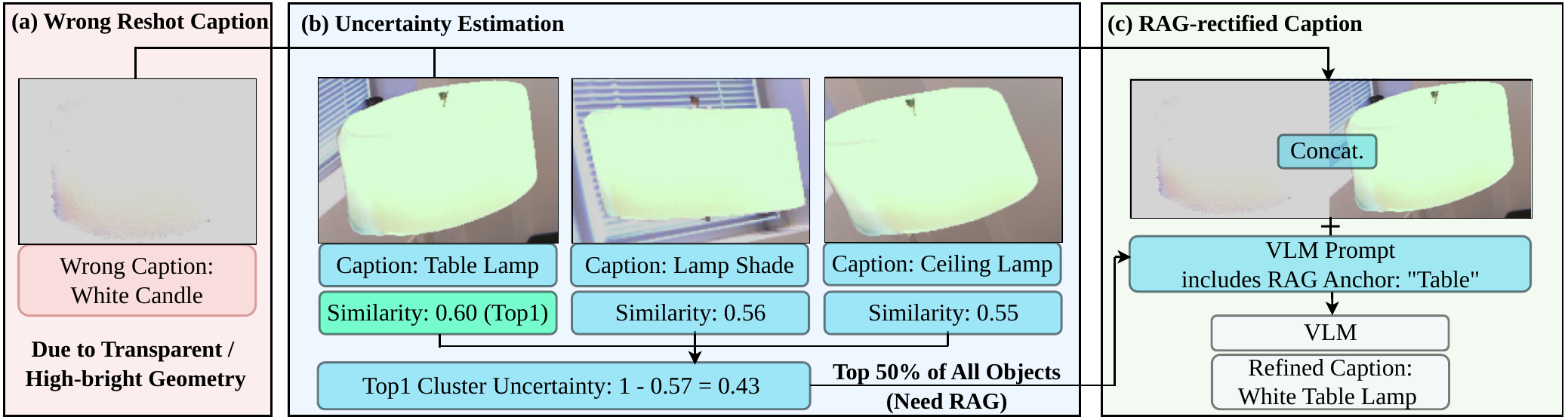}
   \caption{Case study of misleading re-shot recovery.}
   \label{fig:case_study}
   \vspace{-0.5em}
\end{figure}

Fig.~\ref{fig:case_study} shows a bright/transparent table lamp whose re-shot is miscaptioned as ``white candle''. 
Low re-shot/passive agreement triggers RAG:
instead of trusting the re-shot alone, the VLM uses the concatenated
re-shot/top-similarity crop and RAG anchor ``table'' to recover ``white table lamp''.

\vspace{-0.2em}
\subsection{Hyperparameter Ablation Study}
\label{supp:hyperparameter}

To provide a comprehensive understanding of our system's robustness and the trade-offs involved in our algorithmic design, we conduct hyperparameter ablation studies. We structure this analysis around three core parameters: the base downsample voxel size $\delta_{sample} = 0.01$, the merging threshold $\delta_{sim} = 0.45$, and the re-shot scoring weights $\alpha_{prior} = \alpha_{horiz} = 0.2, \alpha_{vis} = 0.6$.

\begin{figure}[ht]
\begin{center}
\includegraphics[width=1\linewidth]{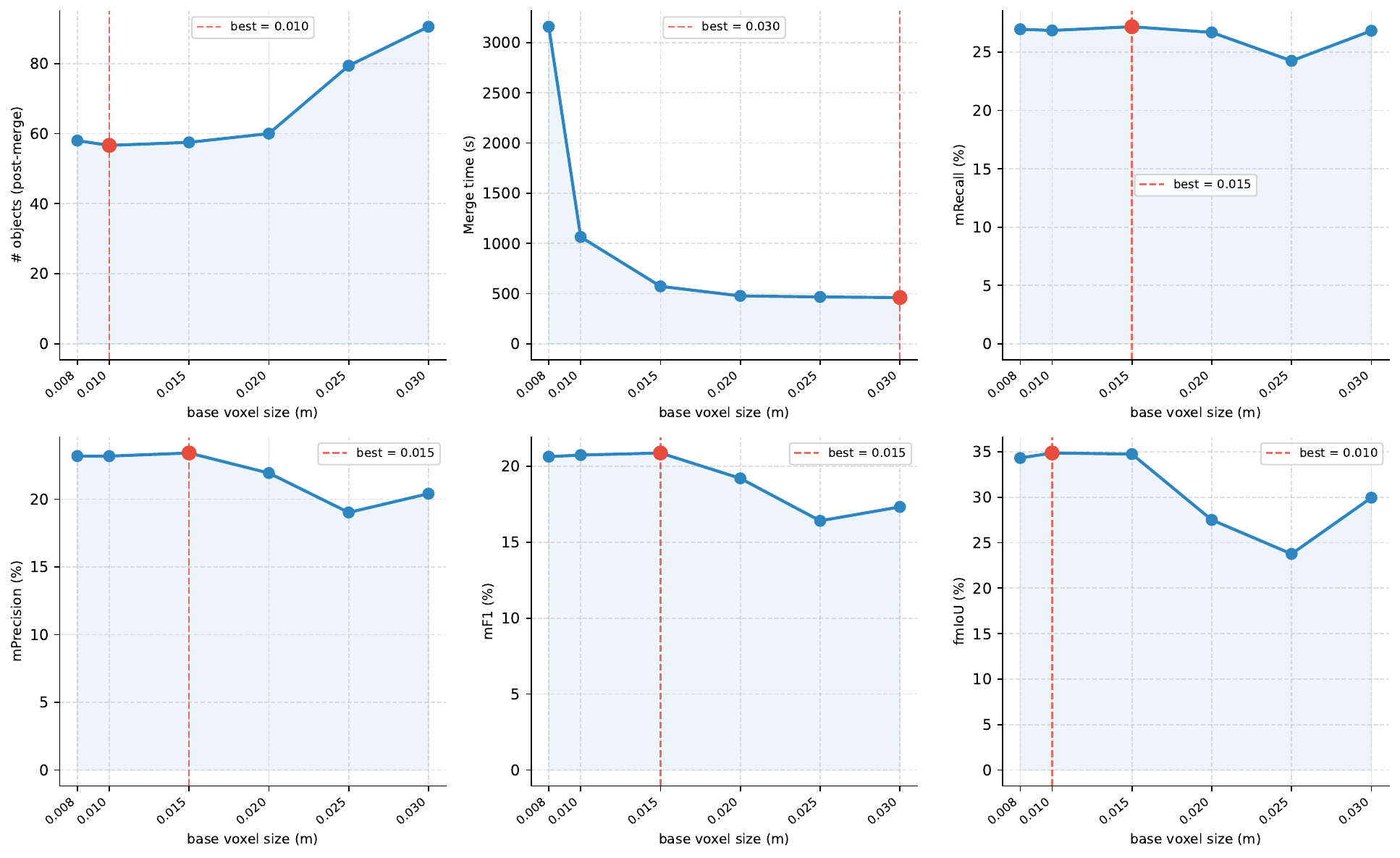}
\end{center}
\caption{
Hyperparameter study for the base downsample voxel size ($\delta^{\text{sample}}$). We evaluate the impact of varying voxel sizes on geometric fragmentation (top-left), computational overhead (top-middle), and semantic reconstruction quality (remaining subplots) averaged across 8 Replica \cite{straub2019replica} scenes. Red markers and dashed lines indicate the peak performance value for each individual metric.
}
\label{fig:downsample}
\end{figure}

\subsubsection{Base Downsample Voxel Size ($\delta^{\text{sample}}$)}

\textbf{Evaluation Protocol:} Experiments are conducted on the Replica \cite{straub2019replica} dataset. 
We follow the evaluation protocol in ConceptGraphs~\cite{gu2024conceptgraphs}. Specifically, the pipeline only executes 2D segmentation and 3D merging. During the merge phase, the semantic embedding of a 3D object is obtained by fusing the CLIP embeddings of its corresponding 2D cropped images. This fused object-level semantic embedding is then directly used to perform 1-NN matching against the per-point Ground Truth (GT) labels (obtained via GradSLAM~\cite{gradslam}).

\textbf{Analysis:} The base downsample voxel size is the cornerstone of our scale-adaptive downsampling strategy, directly dictating point cloud density and computational load. We sweep $\delta^{\text{sample}}$ from $0.008$ m to $0.030$ m (visualized in \cref{fig:downsample}). 
\begin{itemize}
    \item \textbf{Efficiency vs. Fragmentation:} As observed in the top-middle plot, the merge time exhibits an exponential decay. Setting $\delta^{\text{sample}} < 0.010$ m causes the merge time to skyrocket (exceeding 3000 seconds in 8 scenes) due to the immense number of raw points. Conversely, setting the voxel size too large ($\delta^{\text{sample}} > 0.015$ m) artificially inflates the number of post-merge objects (top-left plot, rising from $\sim$60 to nearly 90). This indicates severe over-fragmentation, as the excessively sparse point clouds fail to satisfy the geometric connectivity thresholds during spatial clustering.
    \item \textbf{Semantic Quality:} The metrics clearly demonstrate a ``sweet spot'' between $0.010$ m and $0.015$ m, with fmIoU peaking precisely at $0.010$ m. When the voxel size exceeds $0.015$ m, mPrecision and fmIoU drop precipitously, as coarse voxels destroy the fine-grained geometric boundaries necessary for accurate 3D mask projection.
\end{itemize}
\textbf{Conclusion:} While $\delta^{\text{sample}} = 0.015$ m marginally favors mPrecision and mF1 in the evaluation, we ultimately select $\delta^{\text{sample}} = 0.010$ m as our default configuration. This finer resolution achieves the highest fmIoU and, crucially, preserves denser and smoother geometric surfaces. This high-fidelity geometric preservation is indispensable for our downstream re-shot pipeline.

\subsubsection{Merge Threshold ($\delta_{sim}$)}

\textbf{Evaluation Protocol:} We utilize the exact same truncated evaluation protocol (Seg + Merge only) as described in the previous section. By relying solely on the fused CLIP embeddings for 1-NN matching, we rigorously isolate the impact of the similarity threshold from any downstream VLM generative variations.

\begin{figure}[ht]
\begin{center}
\includegraphics[width=1\linewidth]{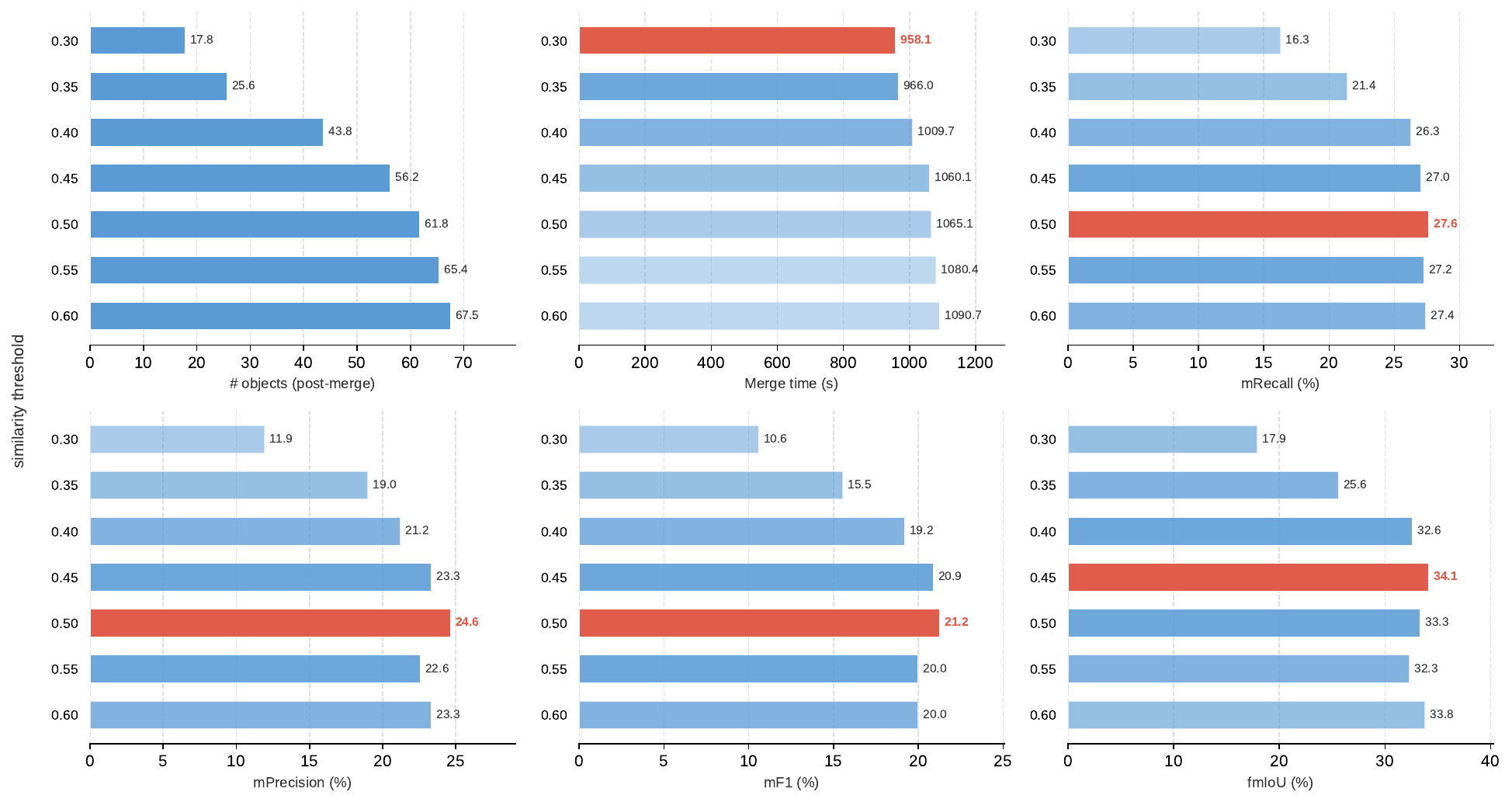}
\end{center}
\caption{
Hyperparameter study for the merging threshold ($\delta_{sim}$). We evaluate the impact of varying similarity thresholds on graph compactness (top-left), processing time (top-middle), and semantic quality (remaining subplots) averaged across 8 Replica \cite{straub2019replica} scenes. Red bars and values indicate the peak performance for each respective metric.
}
\label{fig:merge}
\end{figure}

\textbf{Analysis:} The merging threshold ($\delta_{sim}$) dictates the strictness of the semantic and spatial similarity required to fuse two 3D nodes. A relaxed threshold encourages aggressive merging, while a strict threshold preserves distinct local instances. We sweep $\delta_{sim}$ from 0.30 to 0.60 (visualized in \cref{fig:merge}).
\begin{itemize}
    \item \textbf{Over-merging vs. Fragmentation:} The top-left plot (\# objects) vividly illustrates the core structural trade-off. At low thresholds ($\delta_{sim} \le 0.35$), the system aggressively fuses nodes, resulting in an unnaturally compressed object count ($\sim$17--25). This ``over-merging'' incorrectly blends distinct adjacent objects into indistinguishable semantic blobs, causing a catastrophic collapse across all semantic metrics (e.g., mPrecision plummets to 11.9\% at 0.30). Conversely, excessively high thresholds ($\delta_{sim} \ge 0.55$) trigger overly strict isolation, artificially inflating the object count ($\sim$65--67) and causing unnecessary graph fragmentation without yielding further semantic gains.
    \item \textbf{Semantic Quality:} The metrics establish a clear performance plateau between 0.45 and 0.50. Within this optimal window, the system successfully prevents the fusion of semantically divergent objects while efficiently condensing redundant multi-view observations.
\end{itemize}

\textbf{Conclusion:} While $\delta_{sim} = 0.50$ achieves marginal peak values in mPrecision, mRecall, and mF1, we intentionally select $\delta_{sim} = 0.45$ as our default configuration for two strategic reasons. First, 0.45 attains the absolute highest fmIoU (34.1\%), indicating superior geometric mask alignment. Second, it generates a noticeably more compact scene graph (56.2 objects compared to 61.8 at 0.50). This reduction in redundant nodes directly translates to fewer API calls and lower computational overhead during the following Re-shot and RAG refinement, maintaining high semantic integrity while maximizing system efficiency.

\subsubsection{Re-shot Scoring Weights ($\alpha_{\text{prior}}, \alpha_{\text{horiz}}, \alpha_{\text{vis}}$)}

\textbf{Evaluation Protocol:} Unlike the evaluations above, the effectiveness of the re-shot scoring mechanism is fundamentally tied to the visual quality of the rendered 2D images fed to the downstream VLM. Therefore, for the re-shot scoring weights, we employ a qualitative evaluation protocol to assess viewpoint optimality. To avoid combinatorial explosion and clearly illustrate the physical intuition behind our design, we evaluate discrete extreme configurations to isolate the contribution and vulnerabilities of each scoring heuristic.

\begin{figure}[ht]
\begin{center}
\includegraphics[width=1\linewidth]{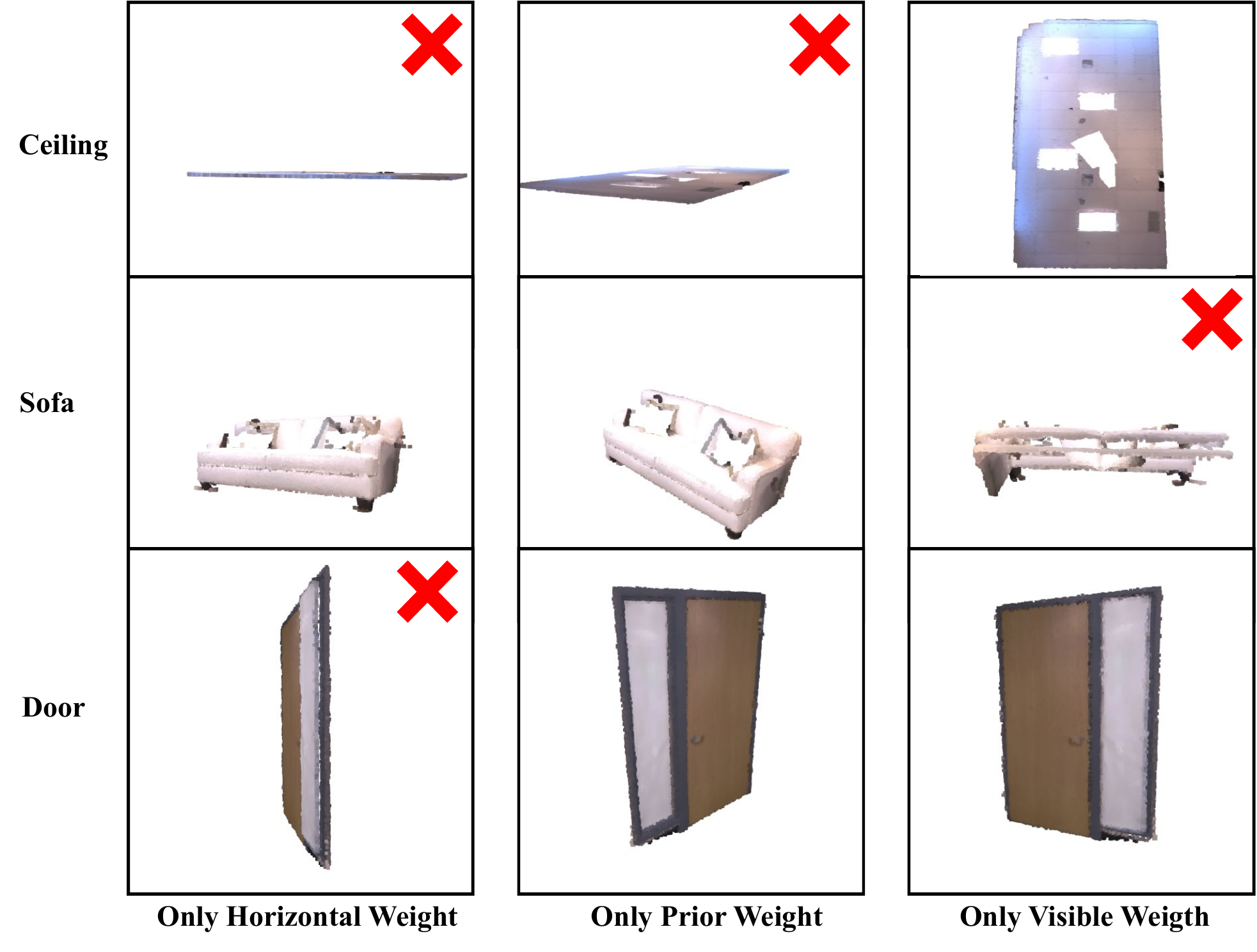}
\end{center}
\caption{
Qualitative ablation of the re-shot scoring weights. We visualize the rendered viewpoints selected by extreme, single-heuristic configurations. Red crosses ($\times$) denote severe failure cases where the selected viewpoint destroys semantic readability. 
}
\label{fig:reshot_weights}
\end{figure}

\textbf{Analysis:} The scoring weights balance three distinct rendering priors: observational consistency ($\alpha_{\text{prior}}$), horizontal viewpoint alignment ($\alpha_{\text{horiz}}$), and information maximization ($\alpha_{\text{vis}}$). We render variants that rely exclusively on a single metric, as visualized in \cref{fig:reshot_weights}.
\begin{itemize}
    \item \textbf{Vulnerability of Only Horizontal Weight ($\alpha_{\text{horiz}} = 1$):} While forcing a horizontal viewpoint works for typical furniture (e.g., the sofa), it fails for planar objects situated along the Z-axis or objects misaligned with the camera's orthogonal axes. As shown in the Ceiling and Door rows of \cref{fig:reshot_weights}, relying solely on the horizontal weight results in meaningless perspectives, collapsing the object's semantic surface into a unidentifiable line.
    \item \textbf{Vulnerability of Only Prior Weight ($\alpha_{\text{prior}} = 1$):} Anchoring purely to the historical observation trajectory ensures consistency but inherits all the flaws of passive recording. If the original trajectory only captured the object from a constrained angle (e.g., glancing at the Ceiling from afar), the re-shot view remains suboptimal, failing to reveal the object's canonical identity.
    \item \textbf{Vulnerability of Only Visible Weight ($\alpha_{\text{vis}} = 1$):} Maximizing visible point count successfully captures large planar surfaces. However, this greedy strategy is geometrically blind to semantic orientations. For objects like the Sofa, it selects the unobserved backside because the rear surface contains a high point count. Rendering these incomplete backsides introduces severe artifacts that critically induce VLM hallucinations.
\end{itemize}

\textbf{Conclusion:} Our multi-factor scoring system acts as a robust optimization. By assigning the highest weight to visibility ($\alpha_{\text{vis}} = 0.6$), we guarantee the generation of dense, high-fidelity geometry. Crucially, the auxiliary weights ($\alpha_{\text{horiz}} = \alpha_{\text{prior}} = 0.2$) act as indispensable regularizers: they penalize backsides and non-canonical views, collectively steering the virtual camera to capture the optimal images for zero-shot VLM captioning.

\subsubsection{RAG Retrieval Settings and Uncertainty Guidance}
\label{supp:rag_setting}

\textbf{Evaluation Protocol:} To evaluate the impact of different retrieval configurations and verify the efficacy of our ``diagnose-then-rectify'' paradigm, we conduct careful investigations on the Replica dataset \cite{straub2019replica}. We examine two critical dimensions of the Object-level RAG module: (i) the number of retrieved semantic anchors ($k \in \{1, 3, 5\}$), and (ii) the necessity of the uncertainty-guided filtering mechanism. In the ``w/o uncert.'' variant, the uncertainty diagnosis phase is entirely bypassed, and contextual refinement is blindly applied to all objects regardless of their underlying semantic ambiguity. The quantitative results are summarized in \cref{tab:rag_ablation}.

\begin{table}[ht]
\centering
\caption{RAG Ablation Studies on the Replica dataset. Ours utilizes top-1 retrieval as the default configuration.}
\label{tab:rag_ablation}
\begin{tabular}{lcc}
\hline
RAG Setting & mIoU (\%) $\uparrow$ & mF1 (\%) $\uparrow$ \\ \hline
\textbf{Ours (RAG top-1)} & \textbf{23.60} & \textbf{30.78} \\
top-3 & 21.32 & 28.26 \\
top-5 & 21.98 & 27.66 \\
w/o uncert. & 20.15 & 26.52 \\ \hline
\end{tabular}
\end{table}

\textbf{Analysis:}
As summarized in Table~\ref{tab:rag_ablation}, the quantitative results demonstrate a clear trade-off between the context size and the refinement precision, yielding several key insights:
\begin{itemize}
    \item \textbf{Retrieval Context Size (top-$k$):} As indicated in \cref{tab:rag_ablation}, introducing more neighbor objects into the RAG context (i.e., scaling from top-1 to top-3 and top-5) consistently degrades both mIoU and mF1. Specifically, top-3 and top-5 settings drop the mF1 from 30.78\% to 28.26\% and 27.66\%, respectively. This performance decay stems from contextual dilution: incorporating distant or less relevant spatial anchors injects irrelevant semantic constraints, which distracts the VLM from accurate localized refinement.
    \item \textbf{Necessity of Uncertainty Guidance:} Bypassing the uncertainty diagnosis mechanism (``w/o uncert.'') leads to the most severe performance drop, with mIoU plunging to 20.15\% and mF1 falling sharply to 26.52\%. This significant degradation underscores that blindly modifying all nodes disrupts already accurate, low-uncertainty semantic anchors, inducing unwanted over-correction and hallucinations.
\end{itemize}

\textbf{Conclusion:} The ablation results firmly justify our design choices. Utilizing a sparse, highly local context ($k=1$) provides the cleanest semantic anchors for spatial reasoning. More importantly, the ``diagnose-then-rectify'' pipeline proves indispensable, ensuring that the refinement process is precisely targeted at ambiguous predictions while preserving reliable nodes.
\subsection{Extended Details on Real-World Robotic Deployment}
\label{supp:real_world_details}

In this section, we provide extended hardware and architectural details regarding the physical deployment of \ours~on our mobile robotic platform, alongside a qualitative robustness analysis and discussion of real-world challenges.

\subsubsection{Sensor Calibration and Temporal Synchronization}
Accurate spatial and temporal alignment between the visual and state-estimation sensors is a fundamental prerequisite for our pipeline. 
\begin{itemize}
    \item \textbf{Extrinsic Calibration:} The rigid spatial transformations between the RealSense D435 RGB-D camera, the CH110 IMU, and the dual Livox MID 360 LiDARs were pre-calibrated by the chassis manufacturer (Agilex). These factory extrinsics proved sufficiently accurate for our scene-level reconstruction, requiring no further manual refinement.
    \item \textbf{Temporal Synchronization:} 
Temporally, incoming RGBD streams are synchronized with a maximum timestamp deviation of 50 ms. To precisely associate these paired visual observations with the continuous LiDAR-based odometry, our system queries the dynamic transformation tree using the exact optical timestamp of the image. Rather than fetching the most recent state, we perform temporal interpolation between discrete odometry states. This synchronization recovers the precise camera pose at the exact moment of image capture, eliminating geometric artifacts caused by temporal misalignment during continuous robot motion.
\end{itemize}

\subsubsection{System Architecture and Compute Distribution}
Given the heavy computational demands of running high-frequency LiDAR-IMU Odometry (LIO) alongside large Vision-Language Models (VLMs), we used an  offline architecture.
During the deployment, the onboard NVIDIA Jetson AGX Xavier acts exclusively as a data acquisition and state-estimation node. 
It records the RealSense RGB-D streams and runs the real-time LIO pipeline. 
The Xavier logs the sensor data alongside the continuously updated coordinate transformations (\texttt{camera\_link} $\rightarrow$ \texttt{odom} $\rightarrow$ \texttt{map}). 

Once the environmental traversal is complete, the recorded data is transferred to a remote Linux server equipped with high-performance GPUs. The computationally intensive stages of our \ours~framework are executed entirely offline. This paradigm ensures smooth real-world teleoperation without overloading the computing power of the mobile chassis.

\subsubsection{Robustness to Real-World Noise}
\label{supp:real_world_robustness}

\begin{figure}[ht]
\begin{center}
\includegraphics[width=1\linewidth]{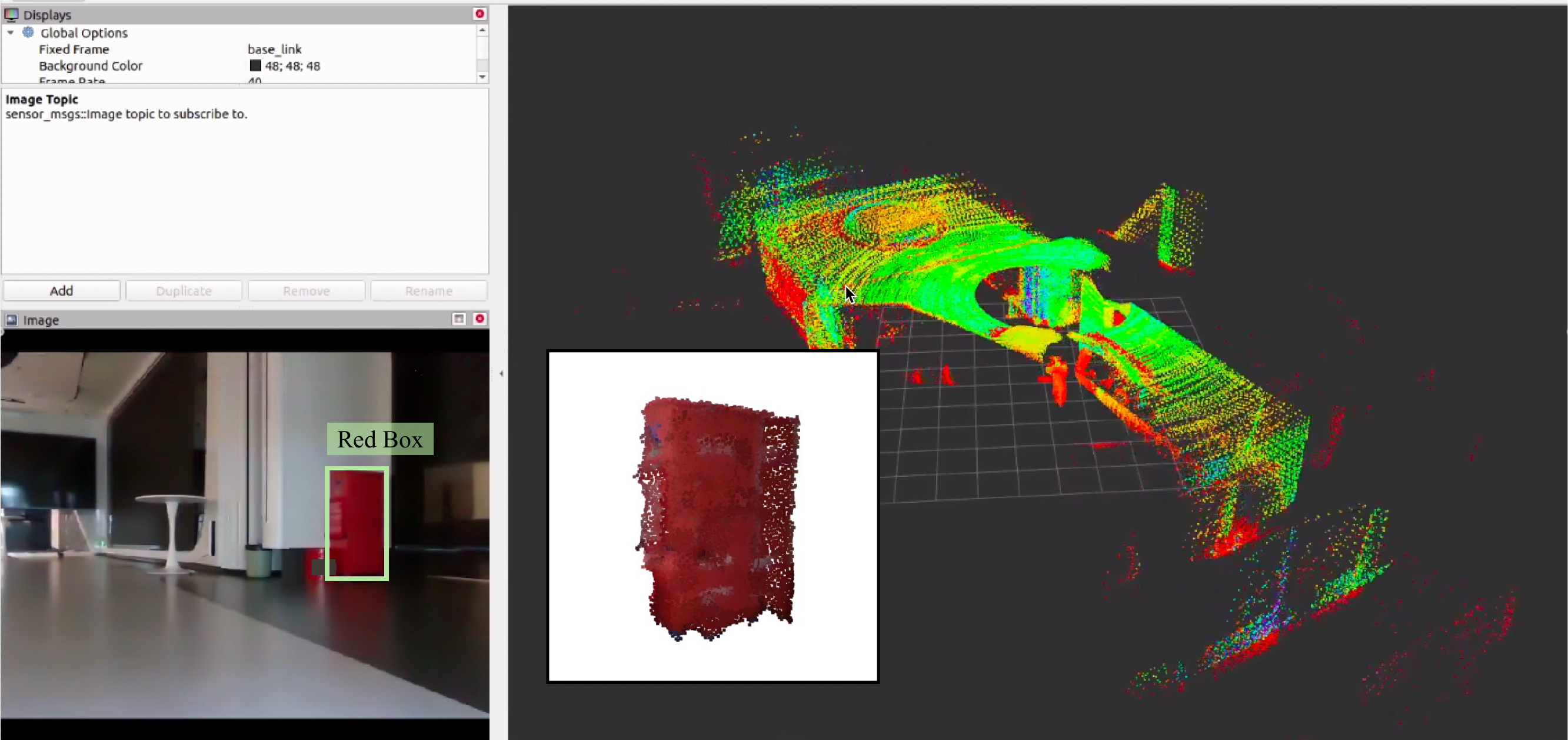}
\end{center}
\caption{
Qualitative visualization of real-world challenges.
Unlike flawless cases of simulated dataset, the real-world case exhibits minor structural distortions due to consumer-grade depth inaccuracies and continuous odometry drift. Despite this degradation, our re-shot and RAG modules successfully extract robust semantic representations.
}
\label{fig:supp_realworld_robustness}
\end{figure}

Transitioning from simulated datasets (e.g., Replica) to physical environments inevitably introduces degradation. To demonstrate the practical resilience of our pipeline against such real-world challenges, we visualize case in \cref{fig:supp_realworld_robustness}.

As illustrated in the inset of \cref{fig:supp_realworld_robustness}, the extracted 3D point cloud of the target object suffers from inherent real-world imperfections. Consumer-grade visual sensors (e.g., RealSense D435) inherently produce noisy depth measurements. Furthermore, even with robust state-estimation (LIO), continuous robot motion introduces minor pose drift over time. Consequently, the reconstructed object exhibits uneven surfaces, artificial thickness, and noisy boundaries.

Crucially, this visualization aligns with the stress-test findings in our main paper. Despite the noticeable degradation, our framework avoids identification failure due to two robust design choices. First, modern VLMs exhibit strong perceptual tolerance to surface noise, provided the semantic structure is effectively captured by our active re-shot mechanism. Second, our pipeline explicitly concatenates the rendered 3D re-shot with the best 2D crop. Therefore, even if the real-world 3D geometry is severely distorted, the uncorrupted 2D crop serves as a dependable visual fallback. This design ensures reliable semantic recognition, proving our system does not require perfect 3D reconstruction.

\subsubsection{Quantitative Real-Robot Evaluation}
\label{supp:realworld_quantitative}

\textbf{Evaluation Protocol:} To systematically quantify the perceptual efficacy of \ours\ in physical deployment settings, we conduct a real-world quantitative evaluation trial on the robotic dataset collected from our indoor facility. We strictly follow the exact same randomized, triple-expert human-evaluation protocol as established in our main text simulation benchmarks. Three independent domain experts evaluate the randomized object nodes to assess the semantic precision of \ours\ against the prominent open-vocabulary baseline, ConceptGraphs~\cite{gu2024conceptgraphs}.

\textbf{Analysis:} The empirical results demonstrate that \ours\ exhibits a compelling advantage in real-world environments. While ConceptGraphs~\cite{gu2024conceptgraphs} suffers heavily from passive-observation noise under consumer-grade sensor constraints, our framework substantially elevates the human-evaluated node precision from \textbf{65.6\%} to \textbf{83.8\%}. This prominent margin (\textbf{+18.2\%}) empirically demonstrates that our active re-shot verification and uncertainty-guided retrieval are exceptionally effective in mitigating physical sensory degradation and non-canonical viewpoint noise during practical robot exploration.

\textbf{Conclusion:} The real-world quantitative validation firmly proves that \ours\ significantly outperforms passive aggregation baselines in suppressing physical noise and hallucinations, establishing a highly robust and precise structured scene representation for practical downstream robotic tasks.

\subsection{Dynamic Downsample}
\label{Dynamic Downsample}
We provide detailed examples of our dynamic downsampling strategy for indoor 3D pointclouds in Table~\ref{tab:dynamic_voxel}. 
The voxel size $\delta_{t,i}^{\text{voxel}}$ for the i-th object in the t-th frame is dynamically adjusted based on its spatial extent:
\begin{equation}
\label{sup_eq:voxel_size}
    \delta_{t,i}^{\text{voxel}} = \delta^{\text{sample}} \cdot \|\text{Bbox}(p_{t,i})\|_2^{1/2},
\end{equation}
where $\delta^{\text{sample}}$ is a base constant and $\|\text{Bbox}(p_{t,i})\|_2$ is the diagonal length of the object's 3D bounding box.

\begin{table}[ht]
\centering
\caption{Examples of dynamic voxel sizes for various indoor objects with a base voxel size of $\delta^{\text{sample}} = 0.01$ m. By utilizing our scale-adaptive downsampling strategy, small objects (e.g., coffee cups and smartphones) are assigned finer voxel resolutions to preserve critical geometric details, whereas expansive structures (e.g., walls and floors) are processed with significantly coarser grids to minimize memory overhead.}
\label{tab:dynamic_voxel}
\resizebox{\textwidth}{!}{
\begin{tabular}{|l|c|c|c|c|}
\hline
\textbf{Object Category} & \textbf{Typical Size (L×W×H)} & \textbf{Diagonal Length} & \textbf{Voxel Size} & \textbf{Reduction Factor} \\
\hline
\multicolumn{5}{|c|}{\textit{Small Objects}} \\
\hline
Coffee cup & $0.08 \times 0.08 \times 0.12$ & 0.165 m & 0.004 m & 2.5× \\
Smartphone & $0.15 \times 0.07 \times 0.008$ & 0.166 m & 0.004 m & 2.5× \\
Computer mouse & $0.12 \times 0.06 \times 0.04$ & 0.14 m & 0.0037 m & 2.7× \\
\hline
\multicolumn{5}{|c|}{\textit{Medium Objects}} \\
\hline
Monitor & $0.55 \times 0.32 \times 0.05$ & 0.638 m & 0.008 m & 1.25× \\
Chair & $0.60 \times 0.55 \times 0.85$ & 1.177 m & 0.011 m & 0.9× \\
Desk & $1.20 \times 0.60 \times 0.75$ & 1.537 m & 0.012 m & 0.8× \\
\hline
\multicolumn{5}{|c|}{\textit{Large Objects}} \\
\hline
Sofa & $2.00 \times 0.90 \times 0.85$ & 2.352 m & 0.015 m & 0.67× \\
Dining table & $2.50 \times 1.20 \times 0.75$ & 2.873 m & 0.017 m & 0.59× \\
Bookshelf & $0.80 \times 0.30 \times 2.20$ & 2.36 m & 0.015 m & 0.67× \\
\hline
\multicolumn{5}{|c|}{\textit{Extra Large Objects}} \\
\hline
Wall & $10.0 \times 0.20 \times 3.0$ & 10.44 m & 0.032 m & 0.3× \\
Floor & $10.0 \times 10.0 \times 0.05$ & 10.00 m & 0.0316 m & 0.3× \\
\hline
\end{tabular}
}
\end{table}

\subsubsection{Robustness Analysis under Varying Point Densities}
\label{supp:density_stress_test}

\textbf{Evaluation Protocol:} To verify the reliability of our visibility scoring metric ($S_{\text{vis}}$) under imperfect or sparse geometric representations, we conduct a density stress test on the Replica dataset \cite{straub2019replica}. We intentionally scale the density of the retained point clouds by $-50\%$, $-25\%$, $+25\%$, and $+50\%$ relative to our default scale-adaptive downsampling configuration. We then evaluate the resulting impact on the final 3D scene graph quality using the mean F1 score (mF1) as the evaluation metric to historical configurations.

\textbf{Analysis:} The empirical evaluations unveil a distinct performance pattern regarding geometric density:
\begin{itemize}
    \item \textbf{Sensitivity to Sparsity:} Decreasing the retained point cloud density by $25\%$ and $50\%$ leads to an mF1 degradation of $-1.45\%$ and $-3.48\%$, respectively, from our default baseline of $30.78\%$. This performance drop indicates that excessively sparse point clouds compromise the approximation accuracy of $S_{\text{vis}}$, leading to suboptimal viewpoint selection.
    \item \textbf{Marginal Benefit of Higher Density:} Conversely, heavily increasing the geometric density by $+25\%$ and $+50\%$ yields negligible performance variations of merely $+0.02\%$ and $+0.07\%$, respectively. This clear plateau implies that once the point cloud density exceeds our default threshold, adding more geometric details provides virtually zero additional semantic or relational information for framework optimization.
\end{itemize}

\textbf{Conclusion:} These stress tests firmly demonstrate that our scale-adaptive downsampling strategy operates precisely near the optimal marginal-benefit regime. It successfully preserves a tight yet sufficient point density required for robust $S_{\text{vis}}$ estimation and viewpoint ranking while aggressively pruning geometric redundancy to minimize system computation and memory overhead.
\subsection{Object Mapping}
\label{appendix:Pseudo Code}
For clarity and reproducibility, we provide the pseudo code of our incremental 3D object mapping algorithm in 
Algorithm~\ref{algorithm:Pseudo Code}. 
This algorithm describes the step-by-step procedure of updating global object list from local object lists of incoming frames, 
including point cloud integration and semantic refinement.
It serves as a concise summary of the mapping pipeline presented in the main paper.

\begin{algorithm}[ht]
\caption{3D Object Fusion with dynamic threshold}
\label{algorithm:Pseudo Code}
\begin{algorithmic}[1]
\REQUIRE Local object list $O_t^{\text{local}}$ for frame $t$, similarity threshold $\delta^{\text{sim}}$, sample threshold $\delta^{\text{sample}}$
\ENSURE Global object list $O_t^{\text{global}}$

\STATE Initialize global object list: $O_1^{\text{global}} = O_1^{\text{local}}$

\FOR{$t = 2$ to $T$}
    \FOR{each local object $o_{t,i}^{\text{local}} = (f_{t,i}^{\text{local}}, p_{t,i}^{\text{local}}) \in O_t^{\text{local}}$}
        \STATE $\text{best\_match} = \text{None}$
        \STATE $\text{max\_similarity} = 0$
        
        \FOR{each global object $o_{t-1,j}^{\text{global}} = (f_{t-1,j}^{\text{global}}, p_{t-1,j}^{\text{global}}) \in O_{t-1}^{\text{global}}$}
            \STATE // Calculate semantic similarity
            \STATE $\theta_{\text{semantic}}(i,j) = (f_{t,i}^{\text{local}})^T f_{t-1,j}^{\text{global}} / 2 + 1/2$
            
            \STATE // Calculate spatial similarity
            \STATE $\delta_{i,j}^{\text{nnratio}} = \delta^{\text{sample}} \cdot (\|\text{Bbox}(p_{t,i}^{\text{local}})\|_2^{1/2} + \|\text{Bbox}(p_{t-1,j}^{\text{global}})\|_2^{1/2}) / 2$
            \STATE $\theta_{\text{spatial}}(i,j) = \text{dnnratio}(p_{t,i}^{\text{local}}, p_{t-1,j}^{\text{global}})$ \COMMENT{Using threshold $\delta_{i,j}^{\text{nnratio}}$}
            
            \STATE // Calculate fusion similarity
            \STATE $\theta(i,j) = \theta_{\text{semantic}}(i,j) + \theta_{\text{spatial}}(i,j)$
            
            \IF{$\theta(i,j) > \text{max\_similarity}$}
                \STATE $\text{max\_similarity} = \theta(i,j)$
                \STATE $\text{best\_match} = j$
            \ENDIF
        \ENDFOR
        
        \IF{$\text{max\_similarity} > \delta^{\text{sim}}$ and $\text{best\_match} \neq \text{None}$}
            \STATE // Fuse with matched global object
            \STATE $j = \text{best\_match}$
            \STATE $n = \text{mapping\_times}(f_{t-1,j}^{\text{global}})$
            \STATE $f_{t,j}^{\text{global}} = (n \cdot f_{t-1,j}^{\text{global}} + f_{t,i}^{\text{local}}) / (n + 1)$
            \STATE $p_{t,j}^{\text{global}} = p_{t-1,j}^{\text{global}} \cup p_{t,i}^{\text{local}}$
            \STATE $o_{t,j}^{\text{global}} = (f_{t,j}^{\text{global}}, p_{t,j}^{\text{global}})$
        \ELSE
            \STATE // Create new global object
            \STATE Add $o_{t,i}^{\text{local}}$ to $O_t^{\text{global}}$ as a new global object
        \ENDIF
    \ENDFOR
\ENDFOR

\RETURN $O_t^{\text{global}} = \{o_{t,1}^{\text{global}}, \ldots, o_{t,N}^{\text{global}}\}$ where $o_{t,i}^{\text{global}} = (f_{t,i}^{\text{global}}, p_{t,i}^{\text{global}})$
\end{algorithmic}
\end{algorithm}
\subsection{Prompt Template}
\label{appendix:Prompt}
In this section, we provide representative examples of the prompt templates.

\subsubsection{Object-Centric Captioning Prompt}
\label{supp:prompt_captioning}
This prompt is utilized to query Vision-Language Models (e.g., GPT-4o \cite{hurst2024gpt} or LLaVA \cite{liu2023llava}) to generate concise semantic descriptions. Within our framework, it is uniformly applied to both the original crops and the newly acquired re-shot optimal views. By explicitly constraining the output format, we ensure that the extracted text can be seamlessly parsed and embedded into the 3D scene graph nodes for subsequent uncertainty estimation and RAG formulation. Crucially, our textual prompt explicitly asks the VLM to describe the ``central object''. This design aligns with our visual preprocessing pipeline: We directly feed the VLM with image crops. Consequently, the target object inherently dominates the geometric center of the input view.

\begin{tcolorbox}[
colback=orange!5!white,
colframe=orange!60!black, 
    title=Object-Centric Captioning Prompt,
    fonttitle=\bfseries,
    arc=1mm,
    breakable
]
\footnotesize
\textbf{Task:} Describe the core identity of the observed object from cropped or re-shot images.

\textbf{Input Prompt:} \\
Briefly describe the central object in the image in a few words. Your response should be the format like this: The object is a xxx xxx.
\end{tcolorbox}

\subsubsection{Background Pruning Prompt}
\label{supp:prompt_background_pruning}

To proactively filter out uninformative structural backgrounds (e.g., walls, floors, or ceilings) and reduce API overhead, we design this early-rejection prompt. Crucially, rather than blindly querying all available crops, we selectively apply this prompt only to the specific historical crop that exhibits the highest semantic similarity to the re-shot anchor. The VLM is instructed to perform strict binary classification and categorization. The strictly formatted output guarantees stable regular expression (Regex) parsing, effectively discarding invalid environmental proposals before they inject semantic noise into the 3D scene graph topology.

\begin{tcolorbox}[
    colback=red!5!white,
    colframe=red!50!black,
    title=Background Pruning Prompt,
    fonttitle=\bfseries,
    arc=1mm,
    breakable
]
\footnotesize
\textbf{Task:} Identify and filter out uninformative structural background elements using the most semantically representative 2D bounding box crop.

\textbf{Input Prompt:} \\
Are there any large areas of walls, floors, or ceilings in the image? Answer yes or no. If yes, choose one type. Your response should be the format like this: yes, xxx. / no.
\end{tcolorbox}

\subsubsection{Object Document Construction Prompt}
\label{supp:prompt_object_document}

This prompt constitutes the core logic of our Object-level Retrieval-Augmented Generation (RAG) module. Its primary objective is to resolve semantic conflicts between the historical multi-view observations (\texttt{image captions}) and the current validation view (\texttt{reshot caption}). To enable the Large Language Model to perform reliable spatial and semantic reasoning, we design a hierarchical rule-based system prompt and employ In-Context Learning (ICL). By providing representative few-shot demonstrations---such as occlusion handling (part-whole relations) and physical dimension comparisons---the LLM learns to systematically select the most accurate object identity and output a structured JSON containing the reasoning path and the final semantic tag.

\begin{tcolorbox}[
    colback=purple!5!white,
    colframe=purple!50!black,
    title=Object Document Construction Prompt (In-Context Learning),
    fonttitle=\bfseries,
    arc=1mm,
    breakable
]
\footnotesize
\textbf{System Prompt:} \\
You are an object identification system for indoor scenes. Your task is to analyze captions and output a structured JSON with reasoning and object tags.

\textbf{Rules (Apply in order of priority):}
\begin{enumerate}
    \item \textbf{Part-whole relation}: If the image captions describe parts of the reshot object, treat the reshot object as correct (likely occlusion).
    \item \textbf{Size comparison}: If the reshot object is larger than the image objects, choose the reshot object. If smaller, choose the image objects.
    \item \textbf{Modifier removal}: If captions describe the same object with different modifiers, remove modifiers and keep only the base object.
    \item \textbf{Semantic similarity}: If captions are semantically similar, pick one representative form.
    \item \textbf{No match}: If none of the above apply, set \texttt{object\_tag} to ``invalid''.
\end{enumerate}

\noindent\rule{\linewidth}{0.4pt}

\textbf{Selected Few-Shot Demonstrations:}

\textit{Demonstration 1: Part-whole relation (Addressing Occlusions)} \\
\textbf{User:} \\
\{ \\
\indent "reshot\_caption": "a bookcase.", \\
\indent "image\_captions": ["some white books.", "a white book.", "some books.", "a bookcase and some books."] \\
\} \\
\textbf{Assistant:} \\
\{ \\
\indent "reason": "image objects(books) are part of the reshot object(bookcase), it may be due to occlusion, so the object is more likely to be a bookcase.", \\
\indent "possible\_tags": ["book", "bookcase"], \\
\indent "object\_tag": "bookcase" \\
\}

\medskip

\noindent \textit{[\dots Additional few-shot demonstrations corresponding to other rules, such as size comparison and modifier removal, are omitted for brevity \dots]}
\end{tcolorbox}

\subsubsection{Object-Level RAG Refinement Prompt}
\label{supp:prompt_rag_refinement}

This prompt executes the core Object-Level Retrieval-Augmented Generation (RAG) mechanism of our framework. For target objects exhibiting high semantic ambiguity, we retrieve a nearby low-uncertainty object to serve as a ``semantic anchor'' based on spatial proximity. By explicitly injecting the validated identity of this anchor into the Vision-Language Model's prompt (e.g., \texttt{\{nearest\_doc\_obj\}}), we provide crucial environmental context. Furthermore, the visual input combines both point cloud projections and RGB crops, seamlessly fusing geometric shapes with appearance textures. This context-guided prompting strategy significantly mitigates hallucinations and enabling it to contextually rectify noisy predictions.

\begin{tcolorbox}[
    colback=teal!5!white,
colframe=teal!60!black,
    title=Object-Level RAG Refinement Prompt,
    fonttitle=\bfseries,
    arc=1mm,
    breakable
]
\footnotesize
\textbf{Task:} Rectify uncertain object predictions by leveraging both multimodal observations (Point Cloud + RGB) and the spatial context of a reliable nearby anchor object.

\textbf{Input Prompt:} \\
The picture is stitched from the point cloud image and RGB image of the same indoor object. And There is a \texttt{\{nearest\_doc\_obj\}} near the object. Briefly tell me what the object in the picture is.
\end{tcolorbox}

\subsubsection{Inter-Object Relationship Generation Prompt}
\label{supp:prompt_edge_generation}

This prompt governs the Edge Caption Generation stage, bridging the gap between isolated semantic nodes to construct a comprehensive 3D Scene Graph. To infer accurate relational edges, we design a prompt that injects both geometric priors (i.e., 3D bounding box centers and extents) and refined semantic identities (\texttt{object\_tag}) into the Large Language Model. By constraining the prediction space to a predefined set of canonical 3D spatial relationships (e.g., support, containment, composition, and proximity) and enforcing strict JSON formatting, we guarantee the topological validity and parsability of the generated scene graph. Few-shot demonstrations are also provided to guide the LLM in applying physical common sense to spatial reasoning.

\begin{tcolorbox}[
colback=green!5!white,
colframe=green!40!black, 
    title=Inter-Object Relationship Generation Prompt,
    fonttitle=\bfseries,
    arc=1mm,
    breakable
]
\footnotesize
\textbf{Task:} Infer the physical and semantic relationship between two 3D objects utilizing their geometric bounding boxes and semantic tags.

\textbf{System Instructions:} \\
The input is a JSON describing two objects ``object1'' and ``object2''. You must output a JSON string (and nothing else). The output JSON must contain exactly two keys: \texttt{object\_relation} (the relationship type) and \texttt{reason} (a brief explanation).

Each of the JSON fields ``object1'' and ``object2'' will have the following fields:
\begin{itemize}
    \setlength{\itemsep}{0pt}
    \setlength{\parskip}{0pt}
    \item \texttt{bbox\_extent}: the 3D bounding box extents of the object
    \item \texttt{bbox\_center}: the 3D bounding box center of the object
    \item \texttt{object\_tag}: an extremely brief description of the object
\end{itemize}

The \texttt{object\_relation} field must be one of the following (verbatim):
\begin{enumerate}
    \setlength{\itemsep}{0pt}
    \setlength{\parskip}{0pt}
    \item ``a on b'' / ``b on a'': if one is commonly placed on top of the other
    \item ``a in b'' / ``b in a'': if one is commonly placed inside the other
    \item ``a part of b'' / ``b part of a'': if one is a physical component of the other
    \item ``near'': if spatially close without clear containment or support
    \item ``none of these'': if none of the above best describe the relationship
\end{enumerate}
\textit{IMPORTANT: Only output a valid JSON string. Do not include extra text, explanation, or formatting outside the JSON.}

\noindent\rule{\linewidth}{0.4pt}

\textbf{Selected Few-Shot Demonstrations:}

\textit{Demonstration 1: Support Relation (on)} \\
\textbf{Input:} \\
\{ \\
\indent "object1": \{"id": 1, "object\_tag": "table"\}, \\
\indent "object2": \{"id": 2, "object\_tag": "floor"\} \\
\} \\
\textbf{Output:} \\
\{ \\
\indent "object\_relation": "a on b", \\
\indent "reason": "A table is typically placed on the floor." \\
\}

\medskip

\noindent \textit{[\dots Additional demonstrations for 'near' and other relations are omitted for brevity \dots]}
\end{tcolorbox}

\end{document}